%% file: main.tex
\documentclass[10pt,journal,twoside,compsoc]{IEEEtran}
\usepackage{times}
\usepackage{epsfig}
\usepackage{graphicx}
\usepackage{amsmath}
\usepackage{amssymb}
\usepackage{bm}
\usepackage{bbm}
\usepackage{graphics}
\usepackage{enumitem} 
\usepackage{url}
\usepackage{hyperref}

\ifCLASSOPTIONcompsoc
  \usepackage[nocompress]{cite}
\else
  \usepackage{cite}
\fi

\input{macros.tex}

\begin{document}
\title{Learning Mesh Representations via Binary Space Partitioning Tree Networks}

\author{Zhiqin~Chen,
        Andrea~Tagliasacchi,
        and~Hao~Zhang
\IEEEcompsocitemizethanks{
\IEEEcompsocthanksitem Z. Chen and H. Zhang are with Simon Fraser University, Burnaby, BC V5A 1S6, Canada. E-mail: {zhiqinc, haoz}@sfu.ca.
\IEEEcompsocthanksitem A. Tagliasacchi is with Google Brain, Toronto, ON M5H 2G4, Canada. E-mail: atagliasacchi@google.com.
}}

\IEEEtitleabstractindextext{
\begin{abstract}
\input{0_abstract}
\end{abstract}

}

\maketitle

\IEEEdisplaynontitleabstractindextext
\IEEEpeerreviewmaketitle

\input{1_intro}

\input{2_related}

\input{3_method}

\input{4_results}

\input{5_future}
\input{6_acks}

\bibliographystyle{IEEEtran}
\bibliography{main}

\newpage
\begin{IEEEbiography}[{\includegraphics[width=1in,height=1.25in,clip,keepaspectratio]{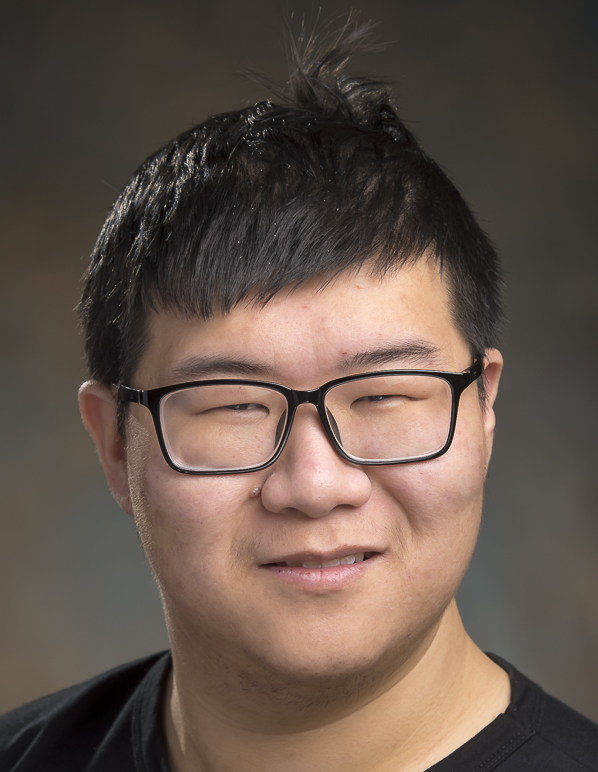}}]{Zhiqin Chen}
is a Ph.D. student at Simon Fraser University, under the supervision of Prof. Hao (Richard) Zhang. He received his Master's degree from Simon Fraser University in 2019, and Bachelor's degree from Shanghai Jiao Tong University in 2017. His research interest is Computer Graphics with specialty in Geometric Modeling and Machine Learning.
\end{IEEEbiography}
\begin{IEEEbiography}[{\includegraphics[width=1in,height=1.25in,clip,keepaspectratio]{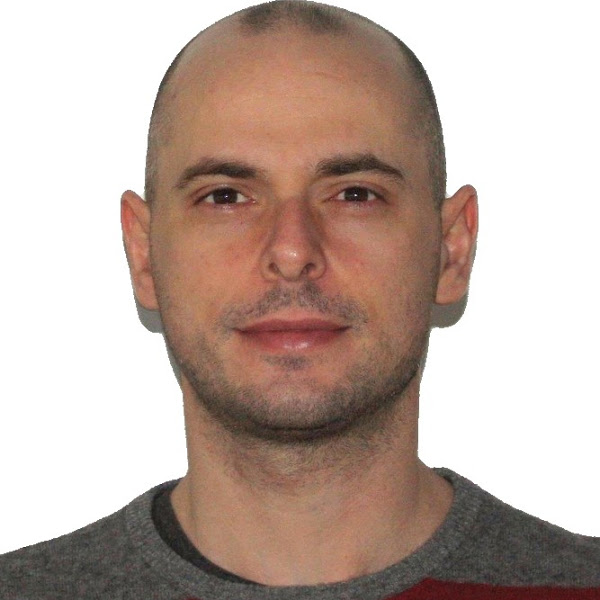}}]{Andrea Tagliasacchi}
is a staff research scientist at Google Brain and an adjunct faculty in the computer science department at the University of Toronto. His research focuses on 3D perception, which lies at the intersection of computer vision, computer graphics and machine learning. In 2018, he was invited to join Google Daydream as a visiting faculty and eventually joined Google full time in 2019. Before joining Google, he was an assistant professor at the University of Victoria (2015-2017), where he held the "Industrial Research Chair in 3D Sensing". His alma mater include EPFL (postdoc) SFU (PhD, NSERC Alexander Graham Bell fellow) and Politecnico di Milano (MSc, gold medalist).
\end{IEEEbiography}
\begin{IEEEbiography}[{\includegraphics[width=1in,height=1.25in,clip,keepaspectratio]{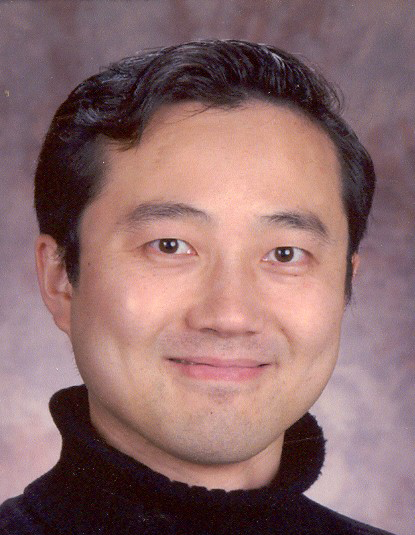}}]{Hao Zhang}
is a professor in the School of Computing Science at SFU and he directs the GrUVi (Graphics U Vision) Lab. He holds a Distinguished SFU Professorship for a five-year term (2020-25). Currently, He is also the Associate Director, Research and Industrial Relations, of his school. His research is in computer graphics and visual computing in general, with special interests in geometric modeling, shape analysis, 3D vision, geometric deep learning, as well as computational design and fabrication.
\end{IEEEbiography}
\vfill

\end{document}

%% file: macros.tex
\usepackage{color}
\definecolor{turquoise}{cmyk}{0.65,0,0.1,0.3}
\definecolor{purple}{rgb}{0.65,0,0.65}
\definecolor{dark_green}{rgb}{0, 0.5, 0}
\definecolor{orange}{rgb}{0.8, 0.6, 0.2}
\definecolor{red}{rgb}{0.8, 0.2, 0.2}
\definecolor{blueish}{rgb}{0.0, 0.7, 1}
\definecolor{light_gray}{rgb}{0.7, 0.7, .7}
\definecolor{pink}{rgb}{1, 0, 1}
\definecolor{dark_red}{rgb}{0.5, 0, 0}

\newcommand{\rz}[1]{{\color{black}#1}}
\newcommand{\zq}[1]{{\color{black}#1}}

\newcommand{\revision}[1]{{\color{black}#1}}

\renewcommand{\paragraph}[1]{{\vspace{1em}\noindent \textbf{#1.}}}

\usepackage{blindtext}

\newcommand{\Figure}[1]{Fig.~\ref{fig:#1}}

\newcommand{\Table}[1]{Table~\ref{table:#1}}
\newcommand{\eq}[1]{(\ref{eq:#1})}

\newcommand{\Section}[1]{Section~\ref{sec:#1}}

\newcommand{\CIRCLE}[1]{\raisebox{.5pt}{\footnotesize \textcircled{\raisebox{-.75pt}{#1}}}}

\newcommand{\object}{\mathrm{G}}
\newcommand{\point}{\mathbf{x}}
\newcommand{\T}{\mathbf{T}}
\newcommand{\W}{\mathbf{W}}
\newcommand{\expect}[2]{\mathbb{E}_{#1}\left[#2\right]}
\DeclareMathOperator*{\argmin}{arg\,min}
\DeclareMathOperator*{\relu}{relu}

\newcommand{\stageone}{+}
\newcommand{\stagetwo}{*}

%% file: 0_abstract.tex
Polygonal meshes are ubiquitous, but have only played a relatively minor role in the deep learning revolution. State-of-the-art neural generative models for 3D shapes learn implicit functions and generate meshes via expensive iso-surfacing. We overcome these challenges by employing a classical spatial data structure from computer graphics, {\em Binary Space Partitioning\/} (BSP), to facilitate 3D learning. The core operation of BSP involves recursive subdivision of 3D space to obtain convex sets. By exploiting this property, we devise BSP-Net, a network that learns to represent a 3D shape via convex decomposition {\em without supervision\/}. The network is trained to {\em reconstruct\/} a shape using a set of convexes obtained from a BSP-tree built over a set of planes, \rz{where the planes and convexes are both defined by learned network weights.} BSP-Net directly outputs polygonal meshes from the inferred convexes. The generated meshes are watertight, {\em compact\/} (i.e., low-poly), and well suited to represent sharp geometry. We show that the reconstruction quality by BSP-Net is competitive with those from state-of-the-art methods while using much fewer primitives. We also explore variations to BSP-Net including using a more generic decoder for reconstruction, more general primitives than planes, as well as training a generative model with variational auto-encoders. Code is available at \href{https://github.com/czq142857/BSP-NET-original}{https://github.com/czq142857/BSP-NET-original}.

%% file: 1_intro.tex
\IEEEraisesectionheading{\section{Introduction}\label{sec:intro}}

\IEEEPARstart{R}{ecently}, there has been an increasing interest in representation learning and generative modeling for 3D shapes. Up to now, deep neural networks for shape analysis and synthesis have been developed mainly for voxel grids~\cite{learningGenerativeEmbeddingCNN,HSP,3DGAN,3DShapeNet}, point clouds~\cite{PCGAN,pointnet,pointnet++,LOGAN,P2PNET,COMPO_NET}, and implicit functions~\cite{imnet,genova_iccv2019,DeepSDF,littwin2019deep,OccNet,DISN,PQ_NET}. As the dominant
3D shape representation for modeling, display, and animation, polygonal meshes
have not figured prominently amid these developments. One of the main reasons is that the non-uniformity and irregularity of triangle tessellations do not naturally support conventional convolution and pooling operations~\cite{hanocka2019}. However, compared to voxels and point clouds, meshes can provide a more seamless and coherent surface representation; they are more controllable, easier to manipulate, and are more {\em compact\/}, attaining higher visual quality using fewer primitives; see~\Figure{teaser}.

\input{fig/teaser}

For display and visualization purposes, the generated voxels, point clouds, and implicits are typically converted into meshes in post-processing, e.g., via iso-surface extraction by Marching Cubes~\cite{marchingcubes}. Few deep networks can generate polygonal meshes directly, and such methods are limited to genus-zero meshes~\cite{Hamu2018,Maron2017,pixel2mesh}, piece-wise genus-zero~\cite{SDM-NET} meshes, meshes sharing the same connectivity~\cite{Gao2017,Tan2018}, or meshes with a very low number of vertices~\cite{scan2mesh}.
Patch-based approaches can generate results which cover a 3D shape with planar polygons~\cite{AOCNN} or curved~\cite{atlasnet} mesh patches, but their visual quality is often tampered by visible seams, incoherent patch connections, and rough surface appearance.
It is difficult to texture or manipulate such mesh outputs.

In this paper, we introduce a neural network which outputs polygonal meshes \textit{natively}. Specifically, 
parameters or weights learned by the network can predict multiple planes to fit the surfaces of a 
3D shape, resulting in a {\em compact\/} and \textit{watertight} mesh; see \Figure{teaser}.
Our network is coined \textit{BSP-Net}, since each facet is associated with a \textit{Binary Space Partitioning} (BSP), 
and the shape is composed by combining these partitions.

\input{fig/outline_tree.tex}

BSP-Net learns an {\em implicit field\/}: given $n$ point coordinates and a shape feature vector as input, the network outputs values indicating whether the points are inside or outside the shape.
The construction of this implicit function is illustrated in \Figure{outline_tree}, and consists of three steps:
\CIRCLE{1} a collection of plane equations implies a collection of $p$ binary partitions of space; see~\Figure{outline_tree}-top;
\CIRCLE{2} an operator~$\mathbf{T}_{p \times c}$ groups these partitions to create a collection of $c$ convex shape primitives/parts;
\CIRCLE{3} finally, the part collection is merged to produce the implicit field of the output shape.

\Figure{outline_net} shows the network architecture of BSP-Net corresponding to these three steps:
\CIRCLE{1} given the feature code, an MLP produces in layer~$L_0$ a matrix $\mathbf{P}_{p \times 4}$ of canonical parameters that define the implicit equations of $p$ planes: $ax + by + cz + d = 0$;
these implicit functions are evaluated on a collection of $n$ point coordinates $\point_{n \times 4}$ in layer $L_1$;
\CIRCLE{2} the operator $\mathbf{T}_{p \times c}$ is a \textit{binary} matrix that enforces a \textit{selective} neuron feed from $L_1$ to the next network layer $L_2$, forming convex parts;
\CIRCLE{3} finally, layer $L_3$ assembles the parts into a shape via either sum or $\min$-pooling.

At inference time, we feed the input to the network to obtain components of the BSP-tree, i.e., leaf nodes (planes $\mathbf{P}$) and connections (binary weights $\mathbf{T}$).
We then apply classic Constructive Solid Geometry (CSG) to extract the explicit polygonal surfaces of the shapes.
The mesh is typically compact, formed by a subset of the $p$ planes directly from the network, leading to a significant speed-up over the previous networks during inference, 
and without the need for expensive iso-surfacing -- current inference time is about 0.5 seconds per generated mesh.
Furthermore, meshes generated by the network are guaranteed to be watertight, possibly with \textit{sharp} features, in contrast to smooth shapes produced by previous implicit decoders~\cite{imnet,DeepSDF,OccNet}.

BSP-Net is trainable and characterized by \textit{interpretable} network parameters defining the hyper-planes and their formation into the reconstructed surface.
Importantly, the network training is \textit{self-supervised} since no ground truth convex shape decompositions are needed.
BSP-Net is trained to \textit{reconstruct} all shapes from the training set using the \textit{same} set of convexes
constructed in layer $L_2$ of the network.
As a result, our network provides a \textit{natural correspondence} between all the shapes at the level of the convexes.
However, BSP-Net does not yet learn semantic parts. Grouping of the convexes into semantic parts can be obtained manually, or learned otherwise as semantic shape segmentation is a well-studied problem.
Such a grouping only needs to be done on each convex once to propagate the semantic understanding to all shapes containing the same semantic parts.

\input{fig/outline_net.tex}

\paragraph{Main contributions}
\begin{itemize}[leftmargin=*,noitemsep]
  \item BSP-Net is the first deep neural network which directly outputs compact and watertight polygonal meshes with 
arbitrary topology and structure variety; it is also the first which can reconstruct and recover {\em sharp geometric features\/}.
  \vspace{3pt}
  \item The learned BSP-tree allows us to easily infer both shape segmentation and part correspondence.
  \vspace{3pt}
  \item By adjusting the encoder of our network, BSP-Net can also be adopted for shape auto-encoding and single-view 3D reconstruction (SVR). 
  \vspace{3pt}
  \item To the best of our knowledge, BSP-Net is among the first to achieve \textit{structured} SVR, reconstructing a \textit{segmented} 
3D shape from a single unstructured object image.
\end{itemize}
\rz{We conduct ablation studies and extensive experiments to evaluate BSP-Net.}
Through evaluations on shape auto-encoding, segmentation, part correspondence, and single-view reconstruction, we demonstrate state-of-the-art performances by our network. Comparisons are made to leading methods on shape decomposition and 3D reconstruction, using conventional distortion metrics, visual similarity, as well as a new metric assessing the capacity of a model in representing sharp features.
In particular, we highlight the favorable fidelity-complexity trade-off exhibited by BSP-Net.

\revision{In addition, we explore several variants to BSP-Net. First, we weaken the interpretability of the network by replacing its selection ($L_2$) and assembly ($L_3$) layers with fully connected layers to obtain more general implicit field decoders. Then, under the same framework, we generalize the primitives from planes to quadratic surfaces. We compare several state-of-the-art implicit decoders including BSP-Net and these variants in terms of a performance and training time trade-off. Finally, we train a preliminary generative model using variational autoencoders.}

%% file: fig/teaser.tex
\begin{figure}[t!]
\begin{center}
\includegraphics[width=1.0\linewidth]{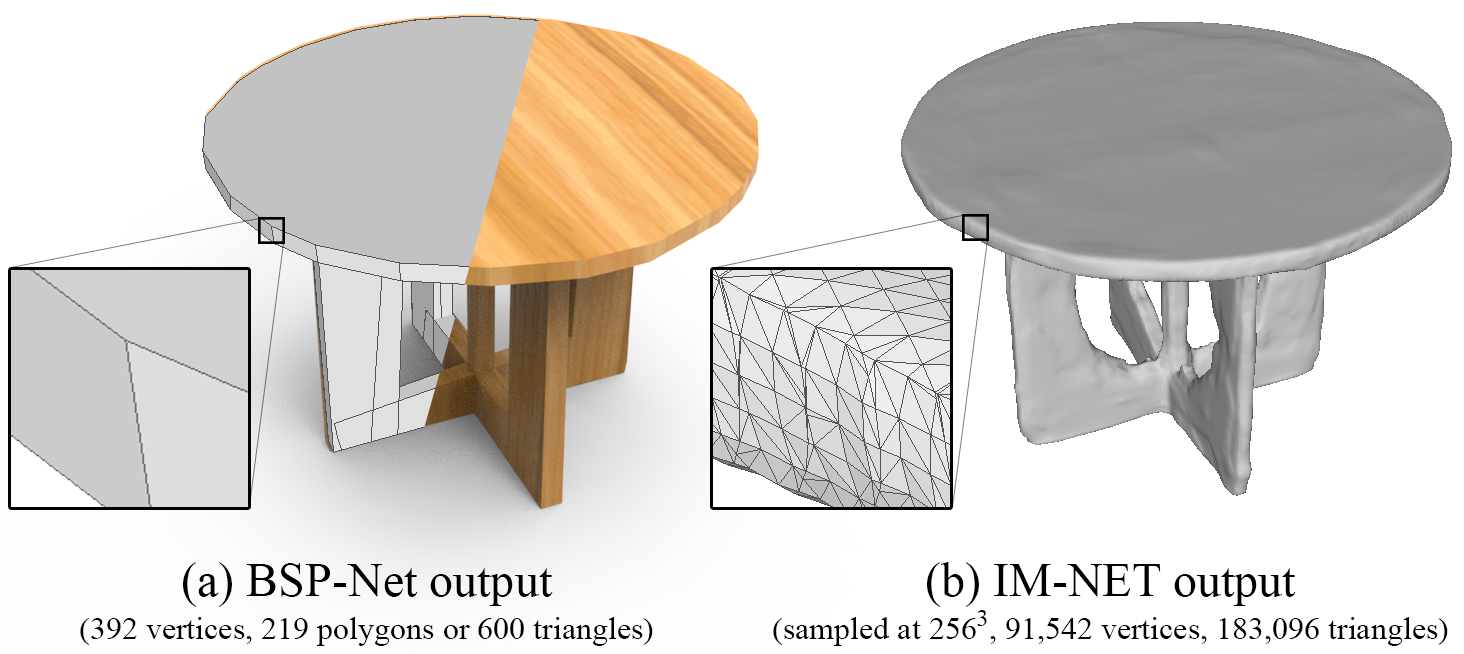}
\end{center}
\caption{
(a) 3D shape auto-encoding 
by our network, BSP-Net, quickly reconstructs a {\em compact\/}, i.e., low-poly, mesh, which can be easily textured. The mesh edges reproduce {\em sharp\/} details in the input~(e.g., edges of the legs), yet still approximate smooth geometry~(e.g., circular table-top).
(b) Current implicit models regress an indicator function, which needs to be iso-surfaced, resulting in over-tessellated meshes which only \textit{approximate} sharp details with smooth surfaces.
}
\label{fig:teaser}
\end{figure}

%% file: fig/outline_tree.tex
\begin{figure}[t!]
\begin{center}
\includegraphics[width=1.0\linewidth]{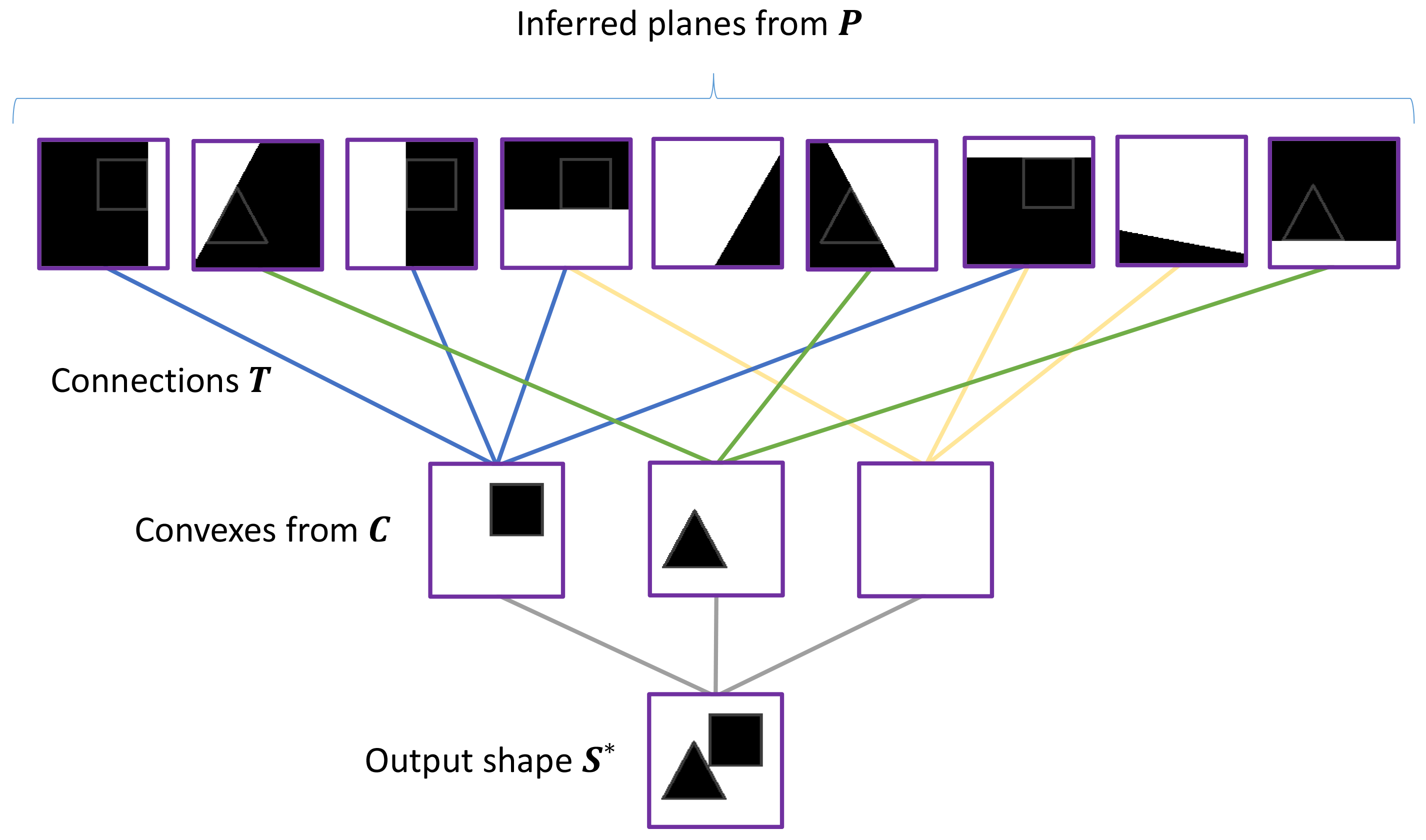}
\end{center}
\caption{
An illustration of the {\em neural\/} BSP-tree for shape construction. \rz{See \Figure{outline_net} for
definitions of $\bm{P}$, $\bm{T}$, $\bm{C}$, and $\bm{S^*}$.}
}
\label{fig:outline_tree}
\end{figure}

%% file: fig/outline_net.tex
\begin{figure}[t!]
\begin{center}
\includegraphics[width=.8\linewidth]{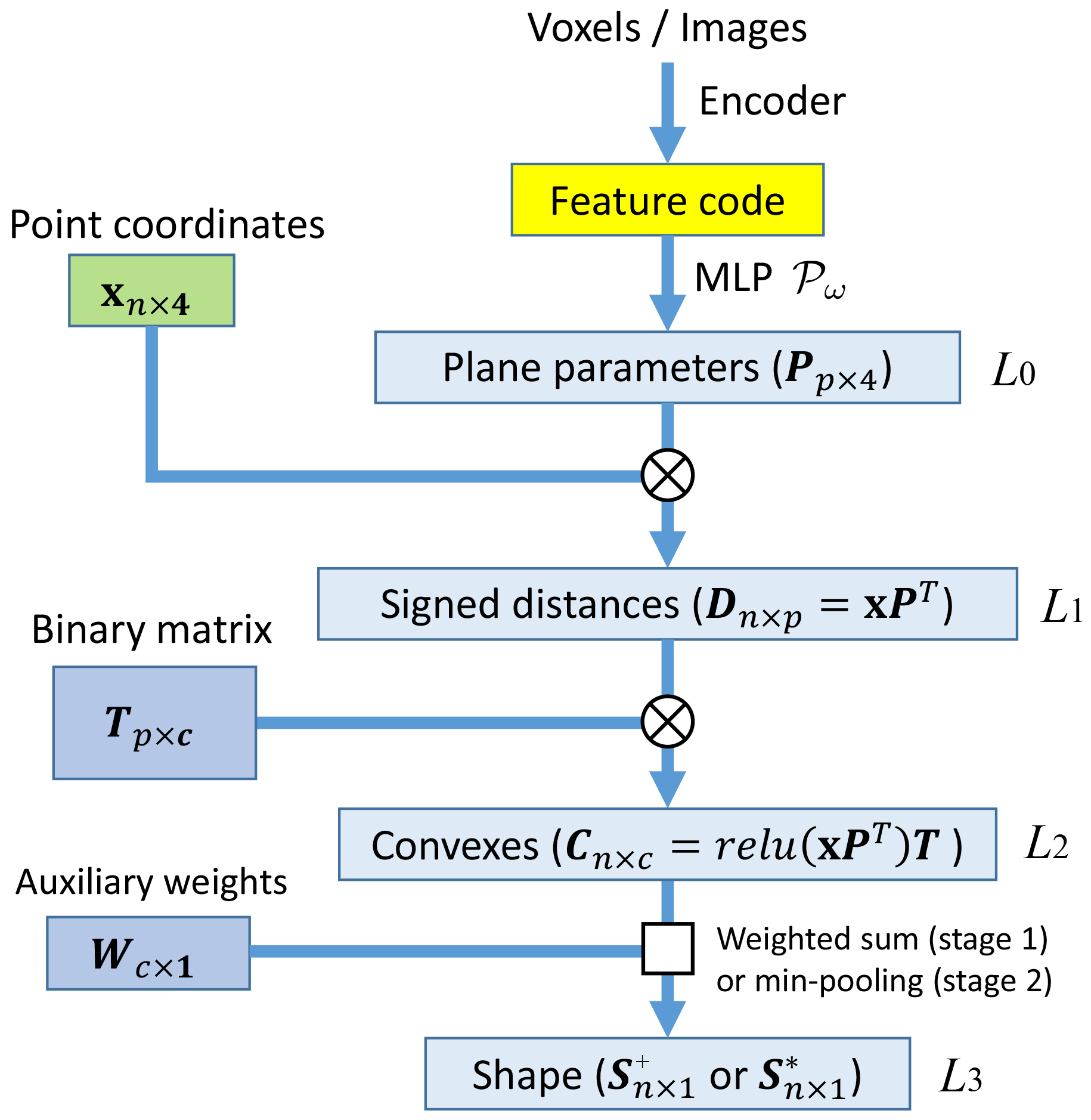}
\end{center}
\caption{
Network architecture of BSP-Net, corresponding to~\Figure{outline_tree}.
}
\label{fig:outline_net}
\end{figure}

%% file: 2_related.tex
\section{Related work}
\label{sec:related}
Large shape collections such as ShapeNet~\cite{chang2015shapenet} and PartNet~\cite{partnet} have spurred the development of learning techniques for 3D data processing.
In this section, we cover representative approaches based on the underlying shape representation learned, with a focus on networks for 3D reconstruction and generative modeling.

\vspace{-5pt}

\paragraph{Grid models}
Early approaches generalized 2D convolutions to 3D~\cite{3DR2N2,learningGenerativeEmbeddingCNN,3DGAN,SVRGAN}, and employed volumetric \textit{grids} to represent shapes in terms of coarse occupancy functions, where a voxel evaluates to zero if it is outside the shape and one otherwise.
Unfortunately, these methods are typically limited to low resolutions of at most $64^3$ due to the cubic growth in memory requirements.
To generate finer results, differentiable marching cubes operations have been proposed~\cite{DeepMarchingCubes}, as well as hierarchical strategies~\cite{HSP,Octnetfusion,octreeNet,OCNN,AOCNN} that alleviate the curse of dimensionality affecting dense volumetric grids.
Another alternative is to use multi-view images~\cite{mutliViewCNN,MVCNN} and geometry images~\cite{DLonGeometryImages,SurfNet}, which allow standard 2D convolution, but such methods are only suitable on the \textit{encoder} side of a network architecture, while we focus on decoders.
Finally, recent methods that perform sparse convolutions~\cite{sparseconv} on voxel grids are similarly limited to encoders.

\vspace{-5pt}

\paragraph{Surface models}
As much of the semantics of 3D models is captured by their {\em surface\/}, the boundary between inside/outside space, a variety of methods have been proposed to represent shape surfaces in a differentiable way.
Amongst these we find a category of techniques pioneered by PointNet~\cite{pointnet} that express surfaces as point clouds~\cite{PCGAN,pointSetGen,PointCloudTree,pointnet,pointnet++,foldingnet,P2PNET}, and techniques pioneered by AtlasNet~\cite{atlasnet} that adopt a 2D-to-3D mapping process~\cite{deeppriors,SurfNet,pixel2mesh,foldingnet}.
An interesting alternative is to consider mesh generation as the process of estimating vertices and their connectivity~\cite{scan2mesh}, but these methods do not guarantee watertight results, and hardly scale beyond a hundred vertices.

\vspace{-5pt}

\paragraph{Implicit models}
A very recent trend has been the modeling of shapes 
as a learnable indicator \textit{function}~\cite{imnet,DeepSDF,OccNet}, rather than a sampling of it, as in the case of voxel methods.
The resulting networks 
treat reconstruction as a \textit{classification} problem, and are universal approximators~\cite{universalapprox} whose reconstruction precision is proportional to the network complexity.
However, at inference time, generating a 3D model still requires the execution of an expensive iso-surfacing operation whose performance scales cubically in the desired resolution.
In contrast, our network directly outputs a \textit{low-poly} approximation of the shape surface.

\vspace{-5pt}

\paragraph{Shape decomposition}
BSP-Net generates meshes using a part-based approach, hence techniques that learn shape decompositions are of particular relevance.
There are methods that decompose shapes as oriented boxes~\cite{tulsiani2017primitives,Im2Struct}, axis aligned gaussians~\cite{genova_iccv2019}, super-quadrics~\cite{paschalidou2019superquadrics}, or a union of indicator functions, in BAE-NET~\cite{chen2019bae_net}. The architecture of our network 
draws inspiration from BAE-NET, which is designed to segment a shape by reconstructing its parts in different branches of the network. For each shape part, 
BAE-NET learns an implicit field by means of a binary classifier. In contrast, BSP-Net explicitly learns a tree structure built on plane subdivisions for bottom-up part
assembly.

Another similar work is CvxNet~\cite{cvxnet}, which decomposes shapes as a collection of convex primitives.
However, BSP-Net differs from CvxNet in several significant ways:
\CIRCLE{1} we target low-poly reconstruction with sharp features, while they target smooth reconstruction;
\CIRCLE{2} their network always outputs $K$ convexes, while the ``right'' number of primitives is learnt automatically in our method;
\CIRCLE{3} our optimization routine is completely different from theirs, as their compositional tree structure is \textit{hard-coded}.

\vspace{-5pt}

\paragraph{Structured models}
There have been recent works on learning structured 3D models, in particular, linear~\cite{3DPRNN,PQ_NET} or
hierarchical~\cite{GRASS,SCORES,SDM-NET,StructureNet,Im2Struct} organization of part bounding boxes. While some methods learn
part geometries separately~\cite{GRASS,StructureNet}, others jointly embed/encode structure and 
geometry~\cite{SAGNET,SDM-NET}. What is common about all of these methods is that they are {\em supervised\/}, and
were trained on shape collections with part segmentations and labels. In contrast, BSP-Net is unsupervised. On the other hand,
our network is not designed to infer shape semantics; it is trained to learn convex decompositions. \rz{To the best of our 
knowledge, before BSP-Net,} there was only one prior work, Im2Struct~\cite{Im2Struct}, which infers part structures from a single-view image. 
However, this work only produces a box arrangement; it does not reconstruct a structured {\em shape\/} like BSP-Net.
\rz{Most recently, concurrent and subsequent works on single-view structured 3D reconstruction have emerged~\cite{PQ_NET,li_eccv2020}.}

\vspace{-5pt}

\paragraph{Binary and capsule networks}
The discrete optimization for the tree structures in BSP-Net bears some resemblance to binary~\cite{BNN} and XNOR~\cite{XNOR-Net} neural networks.
However, only \textit{one} layer of BSP-Net employs binary weights, and our training method differs, as we use a continuous relaxation of the weights in early training.
Further, as our network can be thought of as a simplified scene graph, it holds striking similarities to the principles of capsule networks~\cite{sabour2017}, where low-level capsules (hyperplanes) are aggregated in higher (convexes) and higher (shapes) capsule representations.
Nonetheless, while~\cite{sabour2017} addresses discriminative tasks~(encoder), we focus on generative tasks~(decoder).

%% file: 3_method.tex
\section{Method}
\label{sec:method}
We seek a deep representation of geometry that is both trainable and interpretable. This is
achieved by devising a network architecture that provides a differentiable Binary Space Partitioning tree (BSP-tree) representation\footnote{While typical BSP-trees are binary, we focus on $n$-ary trees, with the ``B'' in BSP referring to binary space partitioning, not the tree structure.}, a classical spatial data structure originated from computer graphics~\cite{schumacher1969study,fuchs1980visible}.
This representation is easily \textit{trainable} as it encodes geometry via implicit functions, and \textit{interpretable} since its outputs are a collection of convex polytopes.
While we generally target 3D geometry, we employ 2D examples to explain the technique without loss of generality. 

We achieve our goal via a network containing three main modules, which act on feature vectors extracted by an encoder corresponding to the type of input data (e.g. the features produced by ResNet for images or 3D CNN for voxels).
In more detail, the first layer \textit{extracts} hyperplanes conditional on the input data, the second layer \textit{groups} hyperplanes in the form of half-spaces to create parts (convexes), and the third layer \textit{assembles} parts together to reconstruct the overall object;~see \Figure{outline_net}.

\vspace{-5pt}

\paragraph{Layer 1: hyperplane extraction}
Given a feature vector $\mathbf{f}$, we apply a multi-layer perceptron $\mathcal{P}$ to obtain plane parameters $\bm{P}_{p \times 4}$, where $p$ is the number of planes -- i.e. $\bm{P} = \mathcal{P}_\omega(\mathbf{f})$.
For any point $\point = (x,y,z,1)$, the product $\bm{D} = \point \bm{P}^T$ is a vector of \textit{signed} distances to each plane -- the $i$th distance is negative if $\point$ is \textit{inside} the $i$th plane, and positive if it is \textit{outside} the $i$th plane, with respect to the plane normal.

\vspace{-5pt}

\paragraph{Layer 2: hyperplane grouping}
To group hyperplanes into geometric primitives, we employ a binary matrix $\bm{T}_{p \times c}$.
Via a max-pooling operation, we aggregate input planes to form a set of $c$ \textit{convex} primitives:
\begin{equation}
C_j^\stagetwo (\point) = \max_i( D_{i} T_{ij} ) 
\quad
\begin{cases}
  <0& \text{inside} \\
  >0& \text{outside}.
\end{cases}
\label{eq:cvxstagetwo}
\end{equation}
Note that during training the gradients would flow through only one (max) of the planes.
Hence, to ease training, we employ a version that replaces max with summation:
\begin{equation}
C_j^\stageone(\point) = \sum_i
  \relu(D_{i}) T_{ij} 
\quad 
\begin{cases}
  = 0& \text{inside} \\
  >0& \text{outside}.
\end{cases}
\end{equation}

\vspace{-5pt}

\paragraph{Layer 3: shape assembly}
This layer groups convexes to create a possibly non-convex output shape via min-pooling:
\begin{equation}
S^\stagetwo (\point) = \min_j( C^\stageone_j (\point) )
\quad
\begin{cases}
  =0& \text{inside} \\
  >0& \text{outside}.
\end{cases}
\label{eq:stage2}
\end{equation}
Note that the use of $C^\stageone$ in the expression above is \textit{intentional}.
We avoid using $C^\stagetwo$ due to the lack of a memory efficient implementation of the operator in TensorFlow~1.

Again, to facilitate learning, we distribute gradients to all convexes by resorting to a (weighted) summation:
\begin{equation}
S^\stageone(\point) {=}\!\left[ \sum_j W_j \! \left[ 1-C^\stageone_j (\point) \right]_{[0,1]} \right]_{[0,1]}
\!\!\!
\begin{cases}
  =1& \!\!\!{\approx}~\text{in} \\
  [0,1)& \!\!\!{\approx}~\text{out},
\end{cases}
\label{eq:stage1}
\end{equation}
where $\W_{c \times 1}$ is a weight vector, and $[ \cdot ]_{[0,1]}$ performs clipping.
During training we will enforce $\W{\approx}\bm{1}$. Note that the inside/outside status here is only \textit{approximate}.
For example, when $\W{=}\bm{1}$, and all $C^\stageone_j{=}0.5$, the point is outside of all convexes, but inside their composition.

\input{fig/2Dresult}

\vspace{-5pt}

\paragraph{Two-stage training}
Losses evaluated on \eq{stage1} will be approximate, but have better gradient than \eq{stage2}.
Hence, we develope a two-stage training scheme where:
\CIRCLE{1} in the \textit{continuous} phase, we try to keep all weights continuous and compute an approximate solution via $S^\stageone(\point)$ -- this would generate an approximate result as can be observed in~\Figure{2Dresult}~(b);
\CIRCLE{2} in the next \textit{discrete} phase, we quantize the weights and use a perfect union to generate accurate results by fine-tuning on $S^\stagetwo(\point)$ -- this creates a much finer reconstruction as illustrated in~\Figure{2Dresult}~(c,d).

Our two-stage training strategy is inspired by classical optimization, where smooth relaxation of integer problems is widely accepted,
and mathematically principled.

\subsection{Training stage 1 -- Continuous}
We initialize $\bm{T}$ and $\bm{W}$ with random zero-mean Gaussian noise having $\sigma\!=\!0.02$, and optimize the network via:
\begin{align}
\argmin_{\omega,\T,\W} \:\: \mathcal{L}^\stageone_\mathrm{rec} + \mathcal{L}^\stageone_\T + \mathcal{L}^\stageone_\W.
\end{align}
Given query points $\point$, our network is trained to match $S(\point)$ to the ground truth indicator function, denoted by $\mathrm{F}(\point|\object)$, 
in a least-squares sense:
\begin{align}
\mathcal{L}^\stageone_\mathrm{rec} = \expect{\point \sim \mathrm{G}} {(S^\stageone(\point) - \mathrm{F}(\point|\mathrm{G}))^2},
\end{align}
where 
$\point{\sim}\object$ indicates a sampling that is specific to the training shape $G$ -- including random samples in the unit box as well as samples near the boundary $\partial \object$; see~\cite{imnet}.
An edge between plane $i$ and convex $j$ is represented by $\T_{ij}{=}1$, and the entry is zero otherwise.
We perform a continuous relaxation of a graph adjacency matrix $\mathbf{T}$, where we require its values to be bounded in the $[0,1]$ range:
\begin{align}
\mathcal{L}^\stageone_\mathrm{T} 
= \sum_{t \in \bm{T}} \max(-t,0)
+ \sum_{t \in \bm{T}} \max(t-1,0).
\end{align}
Note that this is more effective than using a sigmoid activation, as its gradients do not vanish.
Further, we would like $\bm{W}$ to be close to $\bm{1}$ so that the merge operation is a sum:
\begin{align}
\mathcal{L}_\mathrm{W}^\stageone = \sum_{j} | W_j - 1 |.
\end{align}
However, we remind the reader that we initialize with $\bm{W}{\approx}\bm{0}$ to avoid vanishing gradients in early training.

\input{fig/2Dconvex}

\subsection{Training stage 2 -- Discrete}
In the second stage, we first quantize $\bm{T}$ by picking a threshold $\lambda=0.01$ and assign~$t {=} (t{>}\lambda) ? 1{:}0$. Experimentally, we found the values learnt for $\bm{T}$ to be small, which led to our choice of a small threshold.
With the quantized $\T$, we fine-tune the network by:
\begin{align}
\argmin_{\omega} \:\: \mathcal{L}^\stagetwo_\mathrm{recon} + \mathcal{L}^\stagetwo_\text{overlap},
\end{align}
where we ensure that the shape is well reconstructed via:
\begin{align}
\mathcal{L}_\mathrm{recon}^\stagetwo = \expect{\point \sim \object}{\mathrm{F}(\point | G) \cdot \max(S^*(\point),0)} \\ 
+ \expect{\point \sim \object}{(1-\mathrm{F}(\point | G)) \cdot (1-\min(S^*(\point),1))}.
\end{align}
The above loss function pulls $S^\stagetwo(\point)$ towards $0$ if $\point$ should be inside the shape; it pushes $S^\stagetwo(\point)$ beyond $1$ otherwise. 
Optionally, we can also discourage overlaps between the convexes.
We first compute a mask $M$ such that $M(\point, j){=}1$ if $\point$ is in convex $j$ and $\point$ is contained in \textit{more} than one convex, and then evaluate:
\begin{align}
\mathcal{L}_\mathrm{overlap}^\stagetwo = 
-\expect{\point \sim \object}{\expect{j}
{ M(\point, j) C^\stageone_j (\point)}}.
\end{align}

\subsection{Algorithmic and training details}
\label{sec:training_details}

In our 2D experiments, we use $p{=}256$ planes and $c{=}64$ convexes. We use a simple 2D convolutional encoder where each layer downsamples the image by half, and doubles the number of feature channels. We use the centers of all pixels as samples.
In our 3D experiments, we use $p{=}4,096$ planes and $c{=}256$ convexes.
The encoder for \textit{voxels} is a 3D~CNN encoder where each layer downsamples the grid by half, and doubles the number of feature channels. It takes a volume of size $64^3$ as input.
The encoder for \textit{images} is ResNet-18 without pooling layers that receives images of size $128^2$ as input.
All encoders produce feature codes $|\mathbf{f}|{=}256$.
The dense network $\mathcal{P}_\omega$ has widths $\{ 512, 1024, 2048, 4p \}$ where the last layer outputs the plane parameters.

When training the auto-encoder for 3D shapes, we adopt the progressive training from~\cite{imnet}, on points sampled from grids that are increasingly denser ($16^3$, $32^3$, $64^3$).
Note that the hierarchical training is not necessary for convergence, but results in an $\approx 3\times$ speedup in convergence.
In Stage~1, we train the network on $16^3$ grids for 228 epochs with batch size $36$, then $32^3$ for 228 epochs with batch size $36$, then $64^3$ for 228 epochs with batch size $12$.
In Stage~2, we train the network on $64^3$ grids for 228 epochs with batch size $12$.

For single-view reconstruction, we also adopt the training scheme in~\cite{imnet}, i.e., train an auto-encoder first, then only train the image encoder of the SVR model to predict \textit{latent} codes instead of directly predicting the output shapes.
We train the image encoder for 1,000 epochs with batch size~$64$.
We run our experiments on a workstation with an Nvidia GeForce RTX 2080 Ti GPU.
When training the auto-encoder (one model on the 13 ShapeNet categories), Stage~1 takes about ${\approx} 3$ days and Stage~2 takes ${\approx} 2$ days; training the image-encoder requires ${\approx} 1$ day.

\subsection{Variants to BSP-Net}
\label{sec:bsp_var}

\input{fig/outline_netfc}

\rz{
\paragraph{BSP-FC: fully connected layers}
By replacing the binary selection and part assembly layers of the original BSP-Net (\Figure{outline_net}) with multiple {\em fully connected layers\/}, we obtain a general implicit field decoder, which we refer to as BSP-FC, as shown in \Figure{outline_netfc}. With this variation, we
forgo interpretability of the network, as it no longer outputs convexes, but would gain with better performance.}
\zq{
Note that before each fully connected layer, we apply leaky ReLU activation function. In the output layer, we apply a leaky clip $h(x) = \max(\min(x, x \cdot 0.01+0.99), x \cdot 0.01)$, if the desired outputs are binary occupancy values.}

\rz{
\paragraph{BSP-FC-QS: quadratic surface primitives}
Furthermore, instead of using planes for binary space partitioning, we could employ more general primitives such as {\em quadratic surfaces\/}. Introducing quadratics into the BSP-Net framework is straightforward: rather than predicting the four parameters $a,b,c,d$ for the plane $ax+by+cz+d=0$, we predict the ten parameters for a quadratic surface $ax^2+by^2+cz^2+fyz+gzx+hxy+px+qy+rz+d=0$. 

Quadratic primitives can greatly increase representational capacity as it allows more complex structures to be built even in the first layer, in contrast to only planes in the original BSP-Net. We denote a BSP-FC network that predicts quadratic surfaces as BSP-FC-QS.
However, it is not straightforward to extract polygonal meshes from quadratic surfaces. And, our experiments showed that naively adding quadratic primitives to BSP-Net as described above does not bring much improvement. Therefore, we leave further investigation into primitive expansions for future work.}

\zq{
During implementation, we observed that BSP-FC and BSP-FC-QS are prone to overfitting. Therefore, we reduce the network capacity by replacing the MLP $\mathcal{P}_\omega$ with a linear mapping that directly maps the feature code to plane parameters. In our experiments, we show that BSP-FC and BSP-FC-QS can be trained faster than other implicit field decoder structures, while maintaining high performance; see Section~\ref{sec:bspvar}.
}

%% file: fig/2Dresult.tex
\begin{figure*}[t!]
\begin{center}
\includegraphics[width=\linewidth]{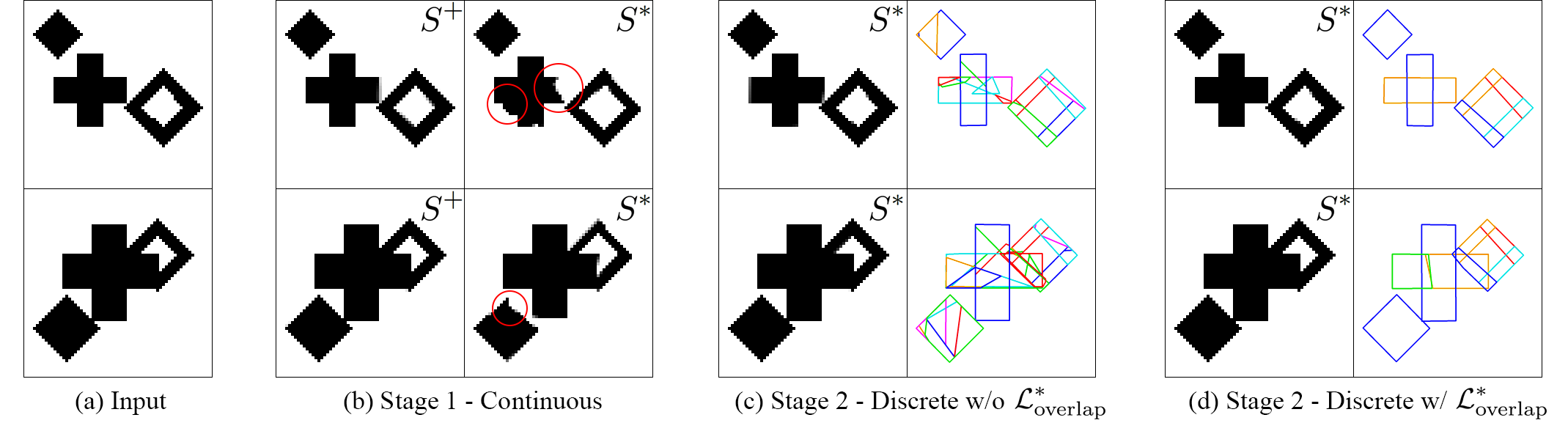}
\end{center}
\vspace{-.5em}
\caption{
\textbf{Evaluation in 2D --}
auto-encoder trained on the synthetic 2D dataset.
We show auto-encoding results and highlight mistakes made in Stage~1 with red circles, which are resolved in Stage~2.
We further show the effect of enabling the (optional) overlap loss.
Notice that in the visualization we use different (possibly repeating) colors to indicate different convexes.
}
\label{fig:2Dresult}
\end{figure*}

%% file: fig/2Dconvex.tex
\begin{figure}[t!]
\begin{center}
\includegraphics[width=1.0\linewidth]{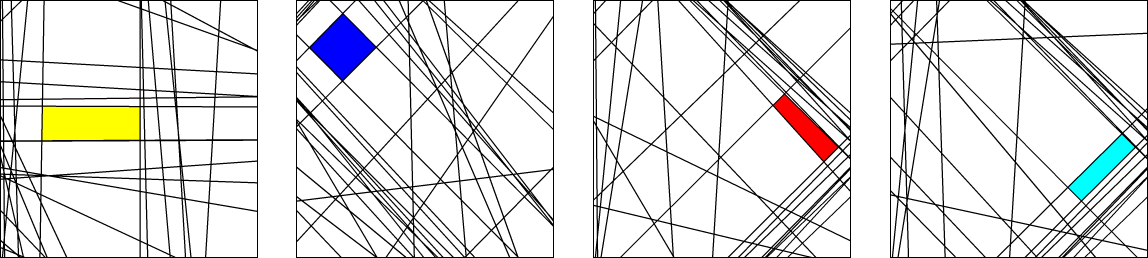}
\end{center}
\vspace{-.5em}
\caption{
\textbf{Examples of $L_2$ output --}
a few convexes from the first shape in \Figure{2Dresult} and the planes used.
Note how many planes are {\em unused\/}.
} 
\label{fig:2Dconvex}
\end{figure}

%% file: fig/outline_netfc.tex
\begin{figure}[t!]
\begin{center}
\includegraphics[width=.8\linewidth]{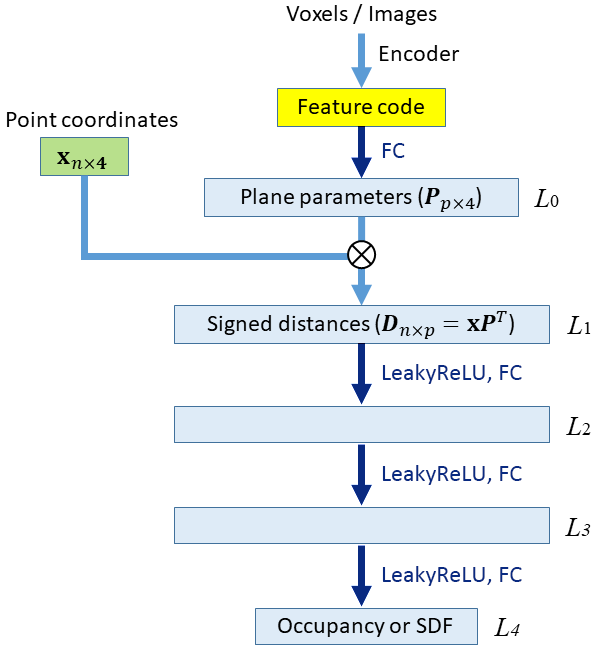}
\end{center}
\caption{
\textbf{The architecture of BSP-FC.} \revision{Compared to the original BSP-Net architecture in~\Figure{outline_net}, we use the same network structure up to layer $L_1$, and then, instead of primitive selection and assembly, we continue with multiple fully-connected (FC) layers to reconstruct the 3D shape. Also, we replace the MLP $\mathcal{P}_\omega$ in BSP-Net by a linear mapping. All new changes are highlighted in dark blue color.}
}
\label{fig:outline_netfc}
\end{figure}

%% file: 4_results.tex
\section{Results and evaluation}
\label{sec:results}
We first study the behavior of BSP-Net on a synthetic 2D shape dataset (\Section{2Dresults}), and then evaluate our auto-encoder (\Section{3DAEresults}), as well as single view reconstruction (\Section{3DSVRresults}) compared to other state-of-the-art methods.
\rz{Ablation studies are conducted in Section~\ref{sec:ablation}, which are followed by results and experiments related to several
variants of BSP-Net (\Section{bspvar}).}

\subsection{Auto-encoding 2D shapes}
\label{sec:2Dresults}
To illustrate how our network works, we created a synthetic 2D dataset. We place a diamond, a cross, and a hollow diamond with varying sizes over $64 \times 64$ images; see \Figure{2Dresult}(a).
The order of the three shapes is \textit{sorted} so that the diamond is always on the left and the hollow diamond is always on the right -- this is to \textit{mimic} the structure of shape datasets such as ShapeNet~\cite{chang2015shapenet}.
After training Stage~1, our network has already achieved a good approximate $S^\stageone$ reconstruction, however, by inspecting $S^\stagetwo$, the output of our inference, we can see there are several imperfections.
After the fine-tuning in Stage~2, our network achieves near perfect reconstructions. 
Finally, the use of overlap losses significantly improves the compactness of representation, reducing the number of convexes per part; see \Figure{2Dresult}(d).

\input{fig/seg_fine}
\input{tab/seg_recon}
\input{tab/seg_iou}

\Figure{2Dconvex} visualizes the planes used to construct the individual convexes -- we visualize planes $i$ in convex $j$ so that $T_{ij}{=}1$ and $P_{i1}^2+P_{i2}^2+P_{i3}^2{>}\varepsilon$ for a small threshold $\varepsilon$ (to ignore planes with near-zero gradients).
Note how BSP-Net creates a natural \textit{semantic} correspondence across inferred convexes.
For example, the hollow diamond in \Figure{2Dresult}(d) is always made of the same four convexes in the same relative positions~--~this is mainly due to the static structure in $\bm{T}$: different shapes need to \textit{share} the same set of convexes and their associated hyper-planes.

\input{fig/seg_compare}

\subsection{Auto-encoding 3D shapes}
\label{sec:3DAEresults}
For 3D shape autoencoding, we compare BSP-Net to several other well-known shape decomposition networks: Volumetric Primitives~(VP)~\cite{tulsiani2017primitives}, Super Quadrics~(SQ)~\cite{paschalidou2019superquadrics}, and Branched Auto Encoders~(BAE)~\cite{chen2019bae_net}.
Note that for the segmentation task, we also evaluate on BAE*, the version of BAE that uses the values of the predicted implicit function, and not just the classification boundaries -- please note that the surface \textit{reconstructed} by BAE and BAE* are \textit{identical}.

Since all these methods target \textit{shape decomposition} tasks, we train \textit{single} class networks, and evaluate segmentation as well as reconstruction performance.
We use the ShapeNet (Part) Dataset~\cite{yi2016scalable}, and focus on five classes: airplane, car, chair, lamp and table.
For the car class, since none of the networks separates surfaces~(as we perform \textit{volumetric} modeling), we reduce the parts from (wheel, body, hood, roof) $\rightarrow$ (wheel, body); and analogously for lamps (base, pole, lampshade, canopy) $\rightarrow$ (base, pole, lampshade) and tables (top, leg, support) $\rightarrow$ (top, leg).

As quantitative metrics for the reconstruction tasks, we report symmetric Chamfer Distance (\textbf{CD}, scaled by ${\times}1000$) and Normal Consistency (\textbf{NC}) computed on $4k$ surface sampled points.
We also report the Light Field Distance (\textbf{LFD})~\cite{LFD} -- the best-known visual similarity metric from computer graphics based on multi-view projections.
For segmentation tasks, we report the typical mean per-label Intersection Over Union (\textbf{IoU}).

\paragraph{Segmentation}
Table~\ref{table:seg_iou} shows the per category segmentation results.
As we have ground truth part labels for the point clouds in the dataset, after training each network, we obtain the part label for each primitive/convex by \textit{voting}: for each point we identify the nearest primitive to it, and then the point will cast a vote for that primitive on the corresponding part label.
Afterwards, for each primitive, we assign to it the part label that has the highest number of votes.
We use 20\% of the dataset for assigning part labels, and we use all the shapes for testing.
At test time, for each point in the \textit{point cloud}, we find its nearest primitive, and assign the part label of the primitive to the point.
In the comparison to BAE, we employ their one-shot training scheme~\cite[Sec.3.1]{chen2019bae_net}.
Note that BAE* is specialized to the segmentation task, while our work mostly targets part-based reconstruction; as such, the IoU performance in \Table{seg_iou} is an \textit{upper bound} of segmentation performance. 

In \Figure{seg_fine}, we show {\em semantic\/} segmentation and part correspondence results that are {\em implied\/} by BSP-Net autoencoding, revealing how individual parts (left/right arm/leg, etc.) are matched.
Note that with BSP-Net, all shapes are corresponded at the primitive (convexes) level.
Thus to reveal shape semantics, we \textit{manually} group convexes belonging to the same semantic part and assign them the same color, where
the color assignment is done on each convex once, and then propagated to all the shapes.

\vspace{-5pt}

\paragraph{Reconstruction comparison}
BSP-Net achieves significantly better reconstruction quality, while maintaining high segmentation accuracy, as demonstrated in \Table{seg_recon} and \Figure{seg_compare}, where we color each \textit{primitive} based on its inferred part label.
BAE-NET is designed for segmentation, thus produces poor-quality part-based 3D reconstructions.
Note how BSP-Net is able to represent complex parts such as the legs of swivel chairs in~\Figure{seg_compare}, while none of the other tested methods could.

\input{tab/svr_numbers}
\input{fig/svr_compare}
\input{tab/svr_poly}
\input{fig/svr_seg}

\subsection{Single view reconstruction (SVR)}
\label{sec:3DSVRresults}
We compare our method with AtlasNet~\cite{atlasnet}, IM-NET~\cite{imnet} and OccNet~\cite{OccNet} on the task of single view reconstruction.
We report quantitative results in Table~\ref{table:svr_numbers} and Table~\ref{table:svr_numbers_avg}, and qualitative results in \Figure{svr_compare}.
We use the $13$ categories in ShapeNet~\cite{chang2015shapenet} that have more than 1,000 shapes each, and the rendered views from 3D-R2N2~\cite{3DR2N2}.
We train one model on all categories, using 80\% of the shapes for training and 20\% for testing, in a similar fashion to AtlasNet~\cite{atlasnet}.
For other methods, we download the \textit{pre-trained} models released by the authors.
Since the pre-trained OccNet~\cite{OccNet} model has a different train-test split than others, we evaluate it on the intersection of the test splits.

\vspace{-5pt}

\paragraph{Edge Chamfer Distance (ECD)}
To measure the capacity of a model to represent \textit{sharp} features, we introduce a new metric. We first compute an ``edge sampling'' of the surface by generating $16k$ points $\mathbf{S}{=}\{\mathbf{s}_i\}$ uniformly distributed on the surface of a model, and then compute sharpness as: 
$\sigma(\mathbf{s}_i) = \min_{j \in \mathcal{N}_\varepsilon(\mathbf{s}_i)} |\mathbf{n}_i \cdot \mathbf{n}_j|,$
where $\mathcal{N}_\varepsilon(\mathbf{s})$ extracts the indices of the samples in $\mathbf{S}$ within distance $\varepsilon$ from $\mathbf{s}$, and $\mathbf{n}$ is the surface normal of a sample.
We set $\varepsilon{=}0.01$, and generate our edge sampling by retaining points such that $\sigma(\mathbf{s}_i){<}0.1$;~see~\Figure{svr_compare}. 
Given two shapes, the ECD between them is nothing but the Chamfer Distance between the corresponding edge samplings. 

\vspace{-5pt}

\paragraph{Analysis}
Our method achieves \textit{comparable} performance to the state-of-the-art in terms of Chamfer Distance.
As for visual quality, our method also \textit{outperforms} most other methods, which is reflected by the superior results in terms of Light Field Distance.
Similarly to \Figure{seg_fine}, we manually color each convex to show part correspondences in \Figure{svr_seg}.
We visualize the triangulations of the output meshes in \Figure{svr_compare}: our method outputs meshes with a smaller number of polygons than state-of-the-art methods.
Note that these methods cannot generate low-poly meshes, and their vertices are always distributed quasi-uniformly.

Finally, note that our method is the \textit{only} one amongst those tested capable of representing sharp edges -- this can be observed quantitatively in terms of Edge Chamfer Distance, where BSP-Net performs much better.
Note that AtlasNet could also generate edges in theory, but the shape is not watertight and the edges are irregular, as it can be seen in the zoom-ins of~\Figure{svr_compare}.
We also analyze these metrics aggregated on the entire testing set in~\Table{svr_numbers_avg}.
In this final analysis, we also include OccNet$_{128}$ and IM-NET$_{256}$, which are the \textit{original} resolutions used by the authors.
Note the average number of \emph{polygons} inferred by our method is $654$ (recall \#polygons $\leq$ \#triangles in polygonal meshes).

\vspace{10pt}

\subsection{Ablation study}
\label{sec:ablation}

\input{tab/ablation_ae}

\input{tab/ablation_svr}

\input{fig/connection}

\rz{

We mainly consider three factors that may influence the performance of BSP-Net: the cutting threshold $\lambda$, hard vs.~soft thresholding, and the overlap loss. Recall that in the second stage, we first quantize $\bm{T}$ by picking a threshold $\lambda=0.01$ and assign~$t {=} (t{>}\lambda) ? 1{:}0$, thus $\lambda$ directly affects how many connections are kept in the BSP-tree. On the other hand, we directly assign binary values to $\bm{T}$ via a \textit{hard} thresholding. One may wonder whether a \textit{soft} thresholding may lead to better performance, by applying a loss term on $\bm{T}$ to push the values towards 0 or 1 gradually and continuously. Therefore, in the second stage, we could remove the hard thresholding and apply:
\begin{align}
\mathcal{L}^\stagetwo_\mathrm{T} 
= \sum_{t \in \bm{T}} [ \mathbbm{1}(t \le \lambda) \cdot |t| + \mathbbm{1}(t > \lambda) \cdot |t-1| ].
\end{align}

In addition, we can apply a post-processing step to reduce the number of convexes to make the output shapes more compact. We first sample a $64^3$ voxel grid for each convex. Note that the union of the voxel grids for all convexes is a $64^3$ voxel grid for the output shape. For each convex, we examine whether removing this convex would change the output shape, i.e., whether the union of the remaining voxel grids is identical to the voxel grid of the output shape. If removing this convex does not affect the output shape, then it is likely that the convex is inside the shape or overlapping with other convexes. Therefore, we will remove such convexes from the final output. This post-processing step is quite effective in our experiments, See \Table{ablation_ae} and \Table{ablation_svr}. CD is sensitive to our post-processing step since removing convexes would cause a density change of the sampled points. LFD is more robust, since we remove convexes that are mostly invisible.

We show the results of our ablation study on shape auto-encoding and SVR in \Table{ablation_ae} and \Table{ablation_svr}, respectively. There are three counterintuitive aspects to note. First, the results on CD and LFD are close, but seemingly random. There is not a clear pattern how $\lambda$ affects the results. This is perhaps due to those evaluation metrics not being robust, or the network being sensitive to initialization. Second, we observe that the overlap loss does not affect the training on 3D shapes as much as it does to that of 2D shapes. In fact, sometimes the overlap loss slightly increases the number of primitives, which may be because we mostly sample points close to shape surface in 3D cases to reduce computational cost, but sample a uniform grid of points in 2D cases. Third, since increasing $\lambda$ reduces the percentage of active connections in the tree structure, one may expect that the number of primitives should decrease. However, they increase when $\lambda$ increases. This can be explained by looking at \Figure{outline_tree}, where the last convex in the middle layer is eliminated. Since the convexes are obtained by intersections of half-spaces, one could imagine that if more connections are added to \Figure{outline_tree}, the convexes would become smaller, even eliminated. In \Figure{connection}, we visualize the distribution of the values in $\bm{T}$ after stage 1 trained with our default settings.

}

\subsection{Variants to BSP-Net}
\label{sec:bspvar}

\input{fig/2Dtoy}

\paragraph{BSP-FC and BSP-FC-QS}
\rz{We compare some of the latest implicit shape decoders~\cite{DeepSDF,imnet,OccNet,littwin2019deep} with BSP-Net and its variants BSP-FC and BSP-FC-QS, as described in \Section{bsp_var}. The comparisons are conducted on shape autoencoding, a task that is universally supported by all of these methods.}

\zq{
\begin{itemize}
\item \textbf{DeepSDF~\cite{DeepSDF} and IM-NET~\cite{imnet}:} 
These networks take a shape feature code and point coordinates as inputs, and pass them to several fully-connected layers, possibly with skip connections. Note that the author of IM-NET~\cite{imnet} released a version that removed all skip connections for faster training, therefore we denote the original version in~\cite{imnet} as IM-NET and the version without skip connections as IM-NET*.
\item \textbf{OccNet~\cite{OccNet}:} 
Occupancy networks use the shape feature code to predict the parameters for conditional batch normalization~\cite{de2017modulating} in the fully-connected layers, and adopt ResNet~\cite{resnet} blocks in their structures. It is the slowest network to train, therefore, we add a smaller version of OccNet by reducing the neurons in each layer by half, denoted as OccNet-/2.
\item \textbf{DeepMETA~\cite{littwin2019deep}:}
Deep Meta Functionals, or DeepMETA, uses the shape feature code to predict the weights of the fully-connected layers. It is the fastest to train, therefore, we add a larger version by increasing the neurons in each layer by 7 times, denoted as DeepMETA-x8.
\item \textbf{BSP-Net \revision{(stage 1)}:}
We compare with BSP-Net trained only on stage 1. Since BSP-Net imposes various constraints to the network inner structures, we expect it to perform worse than most of the other methods. We test two versions of BSP-Net: BSP-Net-4096p with the default settings (4,096 planes), and BSP-Net-2048p with a small number of 2,048 planes.
\item \textbf{BSP-FC:} 
In a standard BSP-FC, the fully-connected layers after $L_1$ have width $\{ 512, 256, 1 \}$. We include a deeper version, BSP-FC-deeper, which has one more fully-connected layer, giving us $\{ 256, 256, 256, 1 \}$ after $L_1$. Similar to BSP-Net, we test the networks with 2,048 or 4,096 planes.
\item \textbf{BSP-FC-QS:} 
The settings are the same with BSP-FC, except that quadratic surfaces have more parameters compared to planes (10 vs. 4), therefore we reduce the number of quadratic surfaces in a network to at most 2,048.
\end{itemize}

\input{tab/bspfc_numbers}

For a fair comparison, we implemented the decoder structures of prior works within our framework, so that the only variable in our comparison experiments is the decoder structure. We apply the same activation function to the output layers and make the networks predict occupancy values. Similar to BSP-Net, we train the networks to perform autoencoding on 13 ShapeNet categories, and apply progressive training on $16^3$, $32^3$ and $64^3$ grids for 100 epochs each, on one Nvidia Tesla V100 GPU. 

Marching Cubes~\cite{marchingcubes} is employed to extract meshes for all the methods at $256^3$ grid resolution, and the autoencoding results on the testing set are reported in \Table{bspfc_numbers}. We also visualize the performance-training time trade-off in \Figure{bspfc_tradeoff}, where we observe that BSP-FC-QS can be trained faster than most other implicit field decoders, while maintaining a high performance.
}

\input{fig/bspfc_tradeoff}

\input{fig/vae_interp}

\paragraph{Generative models using VAEs}
\revision{Finally, we explore generative modeling as a variant to BSP-Net by}
training variational auto-encoders (VAEs)~\cite{VAE} on individual shape categories to 
directly output polygonal meshes. In addition, we enforce that a linear change in the latent feature code corresponds to linear changes of the plane parameters. To this end, we replace the MLP $\mathcal{P}_\omega$, which is applied to the feature code (see \Figure{outline_net}), with a linear mapping, similar to BSP-FC. In \Figure{vae_interp}, we show some latent-space interpolation results generated this way.

\revision{Note that some convex parts may appear or disappear during the interpolation.
We can interpret this behavior with the help of~\Figure{outline_tree}.
One could imagine that if the first plane in \Figure{outline_tree} (left-most in the first row) is moved towards the left, then the first convex in \Figure{outline_tree} (left-most in the second row) will become narrower and narrower, and eventually disappear.
These seemingly discrete changes are due to continuous changes of the planes.}

%% file: fig/seg_fine.tex
\begin{figure}[t!]
\begin{center}
\includegraphics[width=0.99\linewidth]{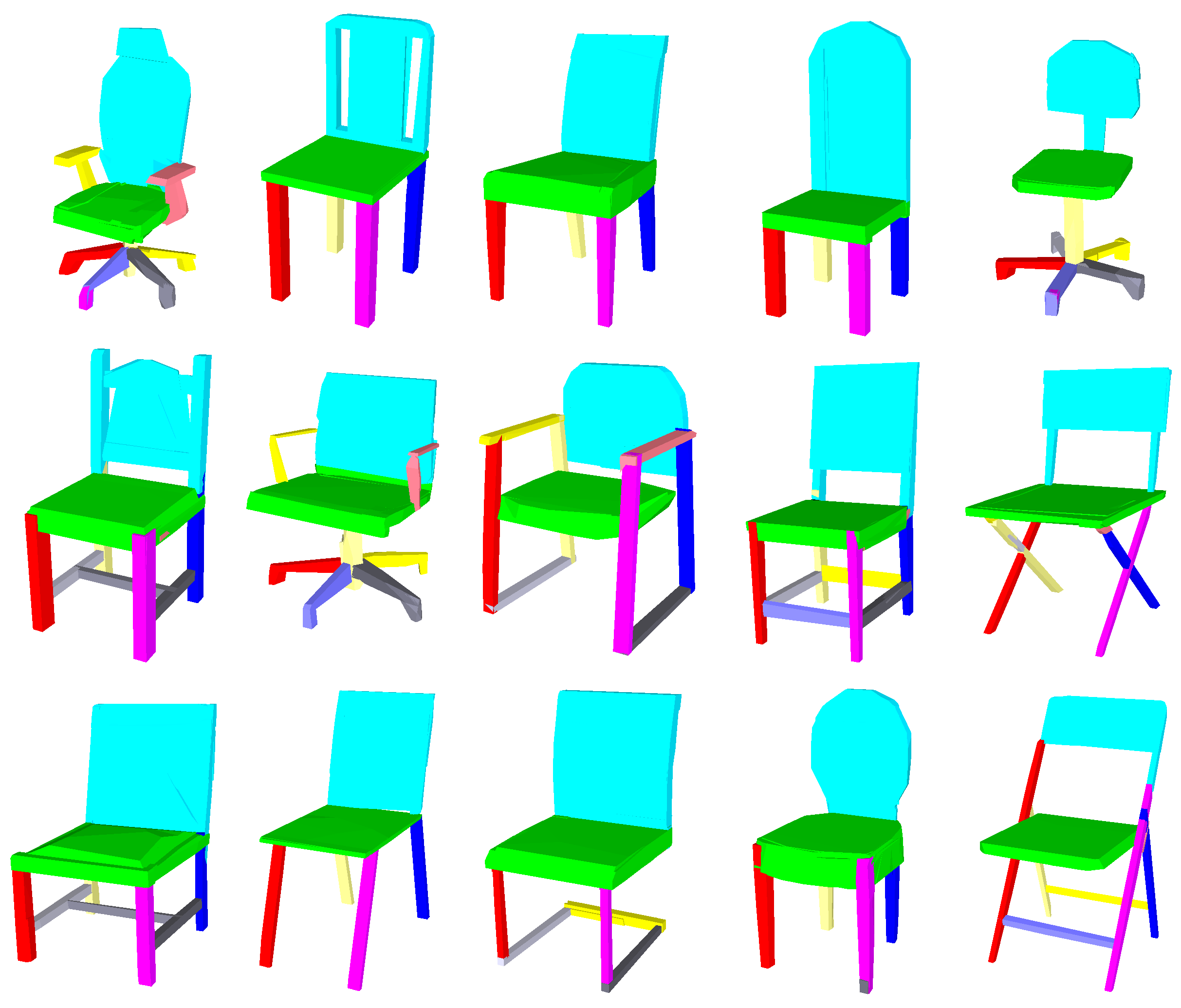}
\end{center}
\vspace{-.5em}
\caption{
\textbf{Segmentation and correspondence -- }
semantics implied from autoencoding by BSP-Net.
Colors shown here are the result of a \textit{manual} grouping of learned convexes.
The color assignment was performed on a few shapes:
once a convex is colored in one shape, we can propagate the color to the other shapes by using the learnt convex ID.
}
\label{fig:seg_fine}
\end{figure}

%% file: tab/seg_recon.tex
\begin{table}[t!]
\caption{
\textbf{Surface Reconstruction}: Comparison for 3D Shape Auto-encoding
\label{table:seg_recon}
}
\vspace{-2em}
\begin{center}
\resizebox{0.7\linewidth}{!}{
\begin{tabular}{l|c|c|c}
\hline
  & CD & NC & LFD \\
\hline\hline
VP~\cite{tulsiani2017primitives}              & 2.259 & 0.683 & 6132.74 \\
SQ~\cite{paschalidou2019superquadrics}        & 1.656 & 0.719 & 5451.44 \\
BAE~\cite{chen2019bae_net}                & 1.592 & 0.777 & 4587.34 \\
Ours                                          & {\bf 0.447} & {\bf 0.858} & {\bf 2019.26} \\
Ours + $\mathcal{L}^\stagetwo_\text{overlap}$                                         & 0.448 & {\bf 0.858} & 2030.35 \\
\hline
\end{tabular}
}
\end{center}
Best results are marked in bold.
\end{table}

%% file: tab/seg_iou.tex
\begin{table}[t!]
\caption{
\textbf{Segmentation}: Comparison in Per-label IoU
\label{table:seg_iou}
}
\vspace{-2em}
\begin{center}
\resizebox{\linewidth}{!}{
\begin{tabular}{l|r|r|r|r|r|r}
\hline
  & plane & car & chair & lamp & table & mean \\
\hline\hline
VP~\cite{tulsiani2017primitives}              & 37.6 & 41.9 & 64.7 & 62.2 & 62.1 & 56.9 \\
SQ~\cite{paschalidou2019superquadrics}        & 48.9 & 49.5 & 65.6 & {\bf 68.3} & 77.7 & 66.2 \\
BAE~\cite{chen2019bae_net}                & 40.6 & 46.9 & 72.3 & 41.6 & 68.2 & 59.8 \\
Ours                                          & 74.2 & 69.5 & 80.9 & 52.3 & {\bf 90.3} & 79.3 \\
Ours + $\mathcal{L}^\stagetwo_\text{overlap}$                                         & {\bf 74.5} & {\bf 69.7} & {\bf 82.1} & 53.4 & {\bf 90.3} & {\bf 79.8} \\
\hline
BAE*~\cite{chen2019bae_net}                & 75.4 & 73.5 & 85.2 & 73.9 & 86.4 & 81.8 \\
\hline
\end{tabular}
}
\end{center}

\end{table}

%% file: fig/seg_compare.tex
\begin{figure}[t!]
\begin{center}
\includegraphics[width=1.0\linewidth]{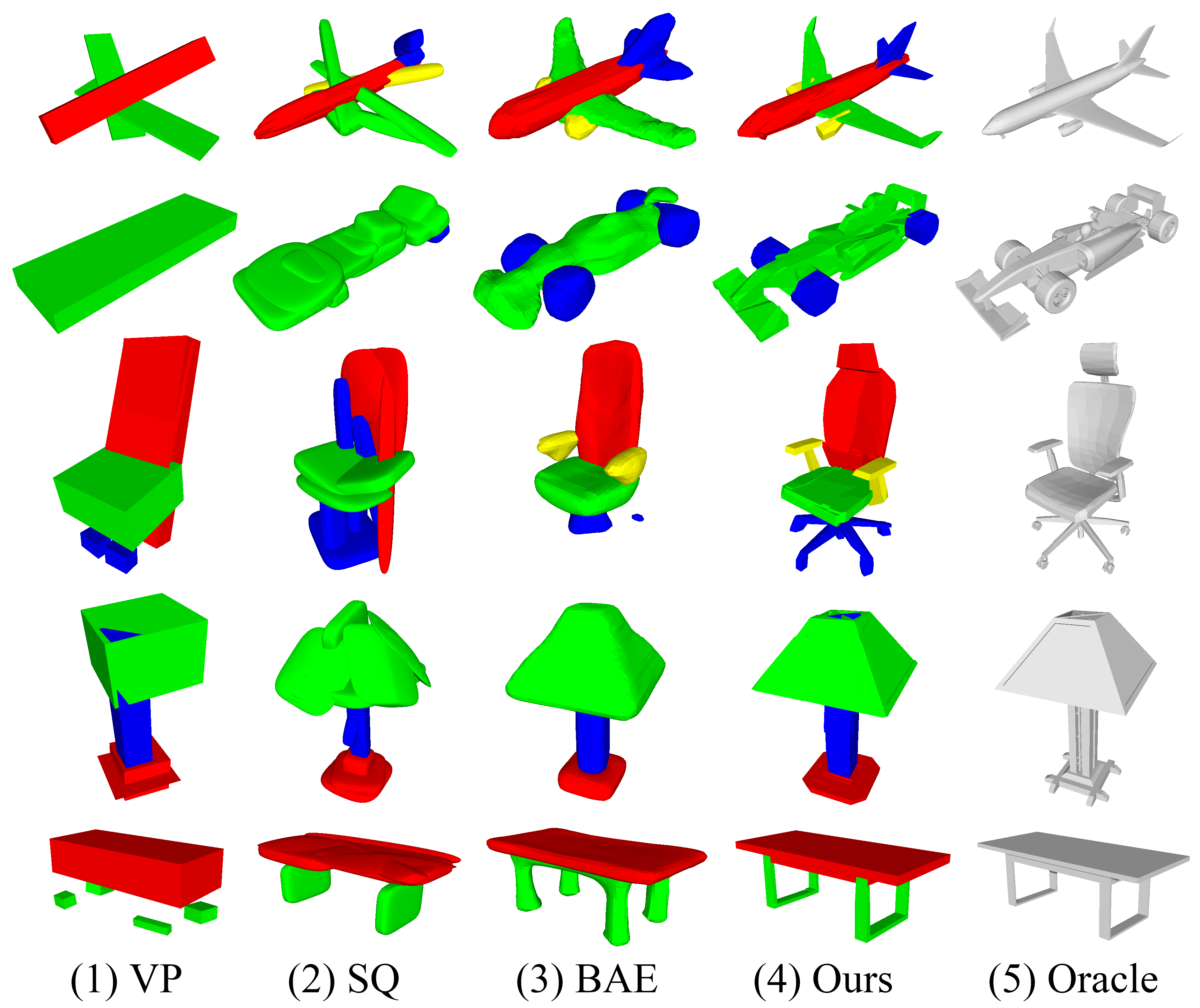}
\end{center}
\vspace{-.5em}
\caption{
\textbf{Segmentation and reconstruction / Qualitative}.
}
\label{fig:seg_compare}
\end{figure}

%% file: tab/svr_numbers.tex
\begin{table*}[t!]
\caption{
\textbf{Single View Reconstruction (SVR)}: Comparison to State-of-the-art Neural Models
\label{table:svr_numbers}
}
\vspace{-2em}
\begin{center}
\resizebox{\linewidth}{!}{
\begin{tabular}{l ||r|r|r|r|r ||r|r|r|r|r  ||r|r|r|r|r }
\hline
  & \multicolumn{5}{|c||}{Chamfer Distance (CD)} & \multicolumn{5}{c||}{Edge Chamfer Distance (ECD)} & \multicolumn{5}{c}{Light Field Distance (LFD)} \\
\hline
  & Atlas0 & Atlas25 & OccNet$_{32}$ & IM-NET$_{32}$ & Ours & Atlas0 & Atlas25 & OccNet$_{32}$ & IM-NET$_{32}$ & Ours & Atlas0 & Atlas25 & OccNet$_{32}$ & IM-NET$_{32}$ & Ours \\
\hline\hline

airplane & 0.587	& {\bf 0.440}	& 1.534	& 2.211	& 0.759	& {\bf 0.396}	& 0.575	& 1.494	& 0.815	& 0.487	& 5129.36	& 4680.37	& 7760.42	& 7581.13	& {\bf 4496.91} \\
bench    & 1.086	& {\bf 0.888}	& 3.220	& 1.933	& 1.226	& 0.658	& 0.857	& 2.131	& 1.400	& {\bf 0.475}	& 4387.28	& 4220.10	& 4922.89	& 4281.18	& {\bf 3380.46} \\
cabinet  & 1.231	& 1.173	& {\bf 1.099}	& 1.902	& 1.188	& 3.676	& 2.821	&10.804	& 9.521	& {\bf 0.435}	& 1369.90	& 1558.45	& 1187.08	& 1347.97	& {\bf 989.12} \\
car      & 0.799	& {\bf 0.688}	& 0.870	& 1.390	& 0.841	& 1.385	& 1.279	& 8.428	& 6.085	& {\bf 0.702}	& 1870.42	& 1754.87	& 1790.00	& 1932.78	& {\bf 1694.81} \\
chair    & 1.629	& {\bf 1.258}	& 1.484	& 1.783	& 1.340	& 1.440	& 1.951	& 4.262	& 3.545	& {\bf 0.872}	& 3993.94	& 3625.23	& 3354.00	& 3473.62	& {\bf 2961.20} \\
display  & 1.516	& {\bf 1.285}	& 2.171	& 2.370	& 1.856	& 2.267	& 2.911	& 6.059	& 5.509	& {\bf 0.697}	& 2940.36	& 3004.44	& 2565.07	& 3232.06	& {\bf 2533.86} \\
lamp     & 3.858	& {\bf 3.248}	&12.528	& 6.387	& 3.480	& 2.458	& 2.690	& 8.510	& 4.308	& {\bf 2.144}	& 7566.25	& 7162.20	& 8038.98	& 6958.52	& {\bf 6726.92} \\
speaker  & 2.328	& {\bf 1.957}	& 2.662	& 3.120	& 2.616	& 9.199	& 5.324	&11.271	& 9.889	& {\bf 1.075}	& 2054.18	& 2075.69	& 2393.50	& 1955.40	& {\bf 1748.26} \\
rifle    & 1.001	& {\bf 0.715}	& 2.015	& 2.052	& 0.888	& 0.288	& 0.318	& 1.463	& 1.882	& {\bf 0.231}	& 6162.03	& 6124.89	& 6615.20	& 6070.86	& {\bf 4741.70} \\
couch    & 1.471	& {\bf 1.233}	& 1.246	& 2.344	& 1.645	& 2.253	& 3.817	&10.179	& 8.531	& {\bf 0.869}	& 2387.09	& 2343.11	& 1956.26	& 2184.28	& {\bf 1880.21} \\
table    & 1.996	& {\bf 1.376}	& 3.734	& 2.778	& 1.643	& 1.122	& 1.716	& 3.900	& 3.097	& {\bf 0.515}	& 3598.59	& 3286.05	& 3371.20	& 3347.12	& {\bf 2627.82} \\
phone    & 1.048	& {\bf 0.975}	& 1.183	& 2.268	& 1.383	&10.459	&11.585	&16.021	&14.684	& {\bf 1.477}	& 1817.61	& 1816.22	& 1995.98	& 1964.46	& {\bf 1555.47} \\
vessel   & 1.179	& {\bf 0.966}	& 1.691	& 2.385	& 1.585	& 0.782	& 0.889	&12.375	& 3.253	& {\bf 0.588}	& 4551.17	& 4430.04	& 5066.99	& 4494.14	& {\bf 3931.73} \\
mean     & 1.487	& {\bf 1.170}	& 2.538	& 2.361	& 1.432	& 1.866	& 2.069	& 6.245	& 4.617	& {\bf 0.743}	& 3644.91	& 3436.14	& 3795.23	& 3700.22	& {\bf 2939.15} \\

\hline
\end{tabular}
}
\end{center}
\textit{Atlas25} denotes AtlasNet with 25 square patches, while \textit{Atlas0} uses a single spherical patch.
Subscripts to OccNet and IM-NET show sampling resolution. For fair comparisons, we use resolution $32^3$ so that OccNet and IM-NET output meshes with comparable number of vertices and faces.
\end{table*}

%% file: fig/svr_compare.tex
\begin{figure}[t!]
\begin{center}
\includegraphics[width=1.0\linewidth]{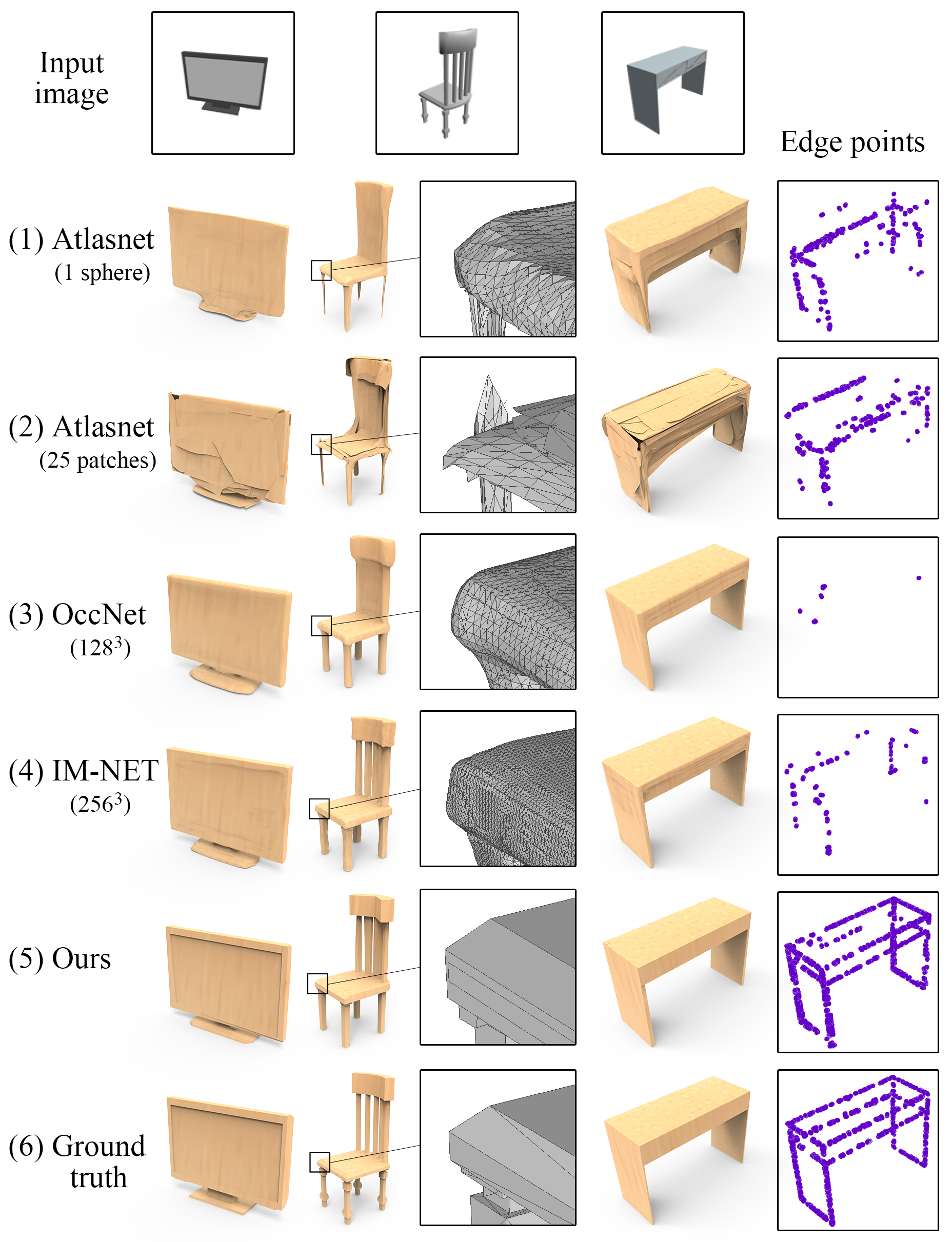}
\end{center}
\vspace{-.5em}
\caption{
\textbf{Single-view 3D reconstruction --} 
comparison to AtlasNet~\cite{atlasnet}, IM-NET~\cite{imnet}, and OccNet~\cite{OccNet}.
Middle column shows mesh tessellations of the reconstruction; last column shows the edge sampling used in the ECD metric.
}
\label{fig:svr_compare}
\end{figure}

%% file: tab/svr_poly.tex
\begin{table}[t!]
\caption{
\textbf{Low-poly Analysis}: Dataset-averaged Metrics for SVR
\label{table:svr_numbers_avg}
}
\vspace{-2em}
\begin{center}
\resizebox{1.0\linewidth}{!}{
\begin{tabular}{l|r|r|r|r|r}
\hline
 & CD & ECD & LFD & \#V & \#F \\
\hline\hline
Atlas0         & 1.487	& 1.866	& 3644.91	& 7446	& 14888 \\
Atlas25        & \textbf{1.170}	& 2.069	& 3436.14	& 2500	& 4050 \\
OccNet$_{32}$  & 2.538	& 6.245	& 3795.23	& 1511	& 3017 \\
OccNet$_{64}$  & 1.950	& 6.654	& 3254.55	& 6756	& 13508 \\
OccNet$_{128}$ & 1.945	& 6.766	& 3224.33	& 27270	& 54538 \\
IM-NET$_{32}$  & 2.361	& 4.617	& 3700.22	& 1204	& 2404 \\
IM-NET$_{64}$  & 1.467	& 4.426	& 2940.56	& 5007	& 10009 \\
IM-NET$_{128}$ & 1.387	& 1.971	& 2810.47	& 20504	& 41005 \\
IM-NET$_{256}$ & 1.371	& 2.273	& \textbf{2804.77}	& 82965	& 165929 \\
Ours	       & 1.432	& \textbf{0.743}	& 2939.15	& \textbf{1073}	& \textbf{1910} \\
\hline
\end{tabular}
}
\end{center}
\#V and \#F are vertex and triangle counts. 
OccNet$_{128}$ and IM-NET$_{256}$ use the original resolutions in the codes provided by the authors.
\end{table}

%% file: fig/svr_seg.tex
\begin{figure}[t!]
\begin{center}
\includegraphics[width=1.0\linewidth]{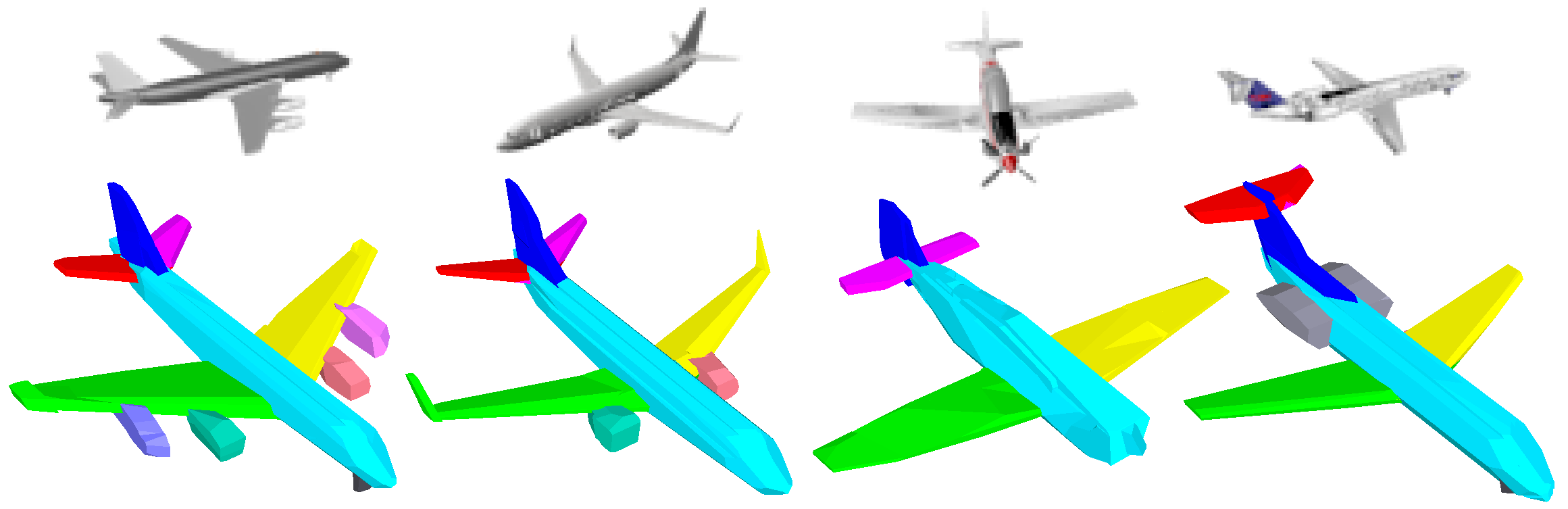}
\end{center}
\vspace{-.5em}
\caption{
\textbf{Structured SVR} by BSP-Net reconstructs each shape with corresponding convexes. Convexes belonging to the same semantic parts are manually grouped and assigned
the same color, resulting in semantic part correspondence.
}
\label{fig:svr_seg}
\end{figure}

%% file: tab/ablation_ae.tex
\begin{table*}[t!]
\zq{
\caption{
Results of Ablation Study on Shape Auto-encoding
\label{table:ablation_ae}
}
}
\vspace{-2em}
\begin{center}
\resizebox{1.0\linewidth}{!}{
\begin{tabular}{l|rr|rr|rr|rr|rr|rr}
\hline
  & \multicolumn{2}{|c}{CD} & \multicolumn{2}{|c}{LFD} & \multicolumn{2}{|c}{\#V} & \multicolumn{2}{|c}{\#F} & \multicolumn{2}{|c}{\#P} & \multicolumn{2}{|c}{\#C} \\
 & Before & After & Before & After & Before & After & Before & After & Before & After & Before & After \\
\hline\hline
Default ($\lambda=0.01$) & 0.743 & 0.764 & 2561.59 & 2563.16 & 1064 & {\bf 508} & 651 & 302 & 1890 & {\bf 920} & 59.3 & 24.1 \\
+ $\mathcal{L}^\stagetwo_\text{overlap}$ & 0.752 & 0.773 & 2581.94 & 2589.53 & 1069 & 513 & 655 & 305 & 1898 & 928 & 59.9 & 24.4 \\
+ $\mathcal{L}^\stagetwo_\mathrm{T}$ & 0.735 & 0.759 & 2533.52 & 2532.89 & 1113 & 517 & 680 & 307 & 1980 & 937 & 61.5 & 24.4 \\
+ $\mathcal{L}^\stagetwo_\mathrm{T}$ + $\mathcal{L}^\stagetwo_\text{overlap}$ & {\bf 0.724} & 0.748 & {\bf 2526.08} & {\bf 2530.77} & 1097 & 522 & 669 & 309 & 1952 & 946 & 60.2 & 24.3 \\
\hline
$\lambda=0.005$ & 0.758 & 0.778 & 2546.90 & 2548.50 & {\bf 1029} & {\bf 508} & {\bf 627} & {\bf 301} & {\bf 1834} & 922 & {\bf 56.2} & {\bf 23.6} \\
$\lambda=0.015$ & 0.745 & 0.767 & 2550.44 & 2557.28 & 1114 & 522 & 680 & 311 & 1982 & 946 & 61.6 & 24.7 \\
$\lambda=0.02$ & 0.766 & 0.792 & 2585.47 & 2584.87 & 1144 & 537 & 699 & 320 & 2035 & 972 & 63.3 & 25.7 \\
$\lambda=0.03$ & 0.747 & 0.772 & 2572.78 & 2575.24 & 1178 & 551 & 722 & 329 & 2092 & 996 & 66.2 & 26.6 \\
$\lambda=0.04$ & {\bf 0.724} & 0.{\bf 746} & 2530.36 & 2531.41 & 1217 & 572 & 741 & 340 & 2169 & 1036 & 66.0 & 26.9 \\
$\lambda=0.05$ & 0.749 & 0.775 & 2564.02 & 2559.96 & 1225 & 580 & 749 & 346 & 2179 & 1049 & 68.0 & 27.8 \\

\hline
\end{tabular}
}
\end{center}
\zq{
\#V, \#F, \#P and \#C denote the number of vertices, triangles, polygons, and convexes, respectively. ``Before'' and ``after'' refer to the post-processing step. \revision{Rows 1-4 show the ablation study on the hard thresholding and the overlap loss; rows 5-10 show the ablation study on parameter $\lambda$.} 
}
\end{table*}

%% file: tab/ablation_svr.tex
\begin{table*}[t!]
\zq{
\caption{
Results of Ablation Study on Single View Reconstruction
\label{table:ablation_svr}
}
}
\vspace{-2em}
\begin{center}
\resizebox{1.0\linewidth}{!}{
\begin{tabular}{l|rr|rr|rr|rr|rr|rr}
\hline
  & \multicolumn{2}{|c}{CD} & \multicolumn{2}{|c}{LFD} & \multicolumn{2}{|c}{\#V} & \multicolumn{2}{|c}{\#F} & \multicolumn{2}{|c}{\#P} & \multicolumn{2}{|c}{\#C} \\
 & Before & After & Before & After & Before & After & Before & After & Before & After & Before & After \\
\hline\hline
Default ($\lambda=0.01$) & 1.432 & {\bf 1.458} & 2939.15 & 2938.29 & 1073 & {\bf 505} & 654 & {\bf 300} & 1910 & {\bf 916} & 58.9 & 23.5 \\
+ $\mathcal{L}^\stagetwo_\text{overlap}$ & 1.447 & 1.480 & 2954.96 & 2960.54 & 1085 & 511 & 663 & 304 & 1930 & 928 & 59.9 & 23.8 \\
+ $\mathcal{L}^\stagetwo_\mathrm{T}$ & 1.493 & 1.528 & 2924.61 & 2925.25 & 1135 & 514 & 691 & 305 & 2023 & 933 & 61.8 & 23.8 \\
+ $\mathcal{L}^\stagetwo_\mathrm{T}$ + $\mathcal{L}^\stagetwo_\text{overlap}$ & 1.453 & 1.482 & {\bf 2913.36} & {\bf 2916.73} & 1117 & 520 & 680 & 307 & 1993 & 944 & 60.4 & 23.8 \\
\hline
$\lambda=0.005$ & 1.478 & 1.509 & 2935.79 & 2942.46 & {\bf 1052} & 508 & {\bf 639} & {\bf 300} & {\bf 1878} & 923 & {\bf 56.4} & {\bf 23.1} \\
$\lambda=0.015$ & 1.492 & 1.522 & 2938.21 & 2946.69 & 1141 & 521 & 695 & 309 & 2034 & 946 & 62.0 & 24.1 \\
$\lambda=0.02$ & 1.507 & 1.541 & 2963.59 & 2961.63 & 1160 & 533 & 707 & 317 & 2066 & 966 & 63.3 & 25.2 \\
$\lambda=0.03$ & {\bf 1.431} & 1.467 & 2943.89 & 2947.45 & 1197 & 549 & 731 & 327 & 2130 & 994 & 66.0 & 25.9 \\
$\lambda=0.04$ & 1.509 & 1.538 & 2936.90 & 2940.44 & 1240 & 569 & 753 & 337 & 2214 & 1033 & 66.5 & 26.3 \\
$\lambda=0.05$ & 1.446 & 1.479 & 2927.04 & 2925.13 & 1257 & 582 & 766 & 346 & 2240 & 1054 & 68.6 & 27.4 \\

\hline
\end{tabular}
}
\end{center}

\end{table*}

%% file: fig/connection.tex
\begin{figure}[t!]
\begin{center}
\includegraphics[width=0.9\linewidth]{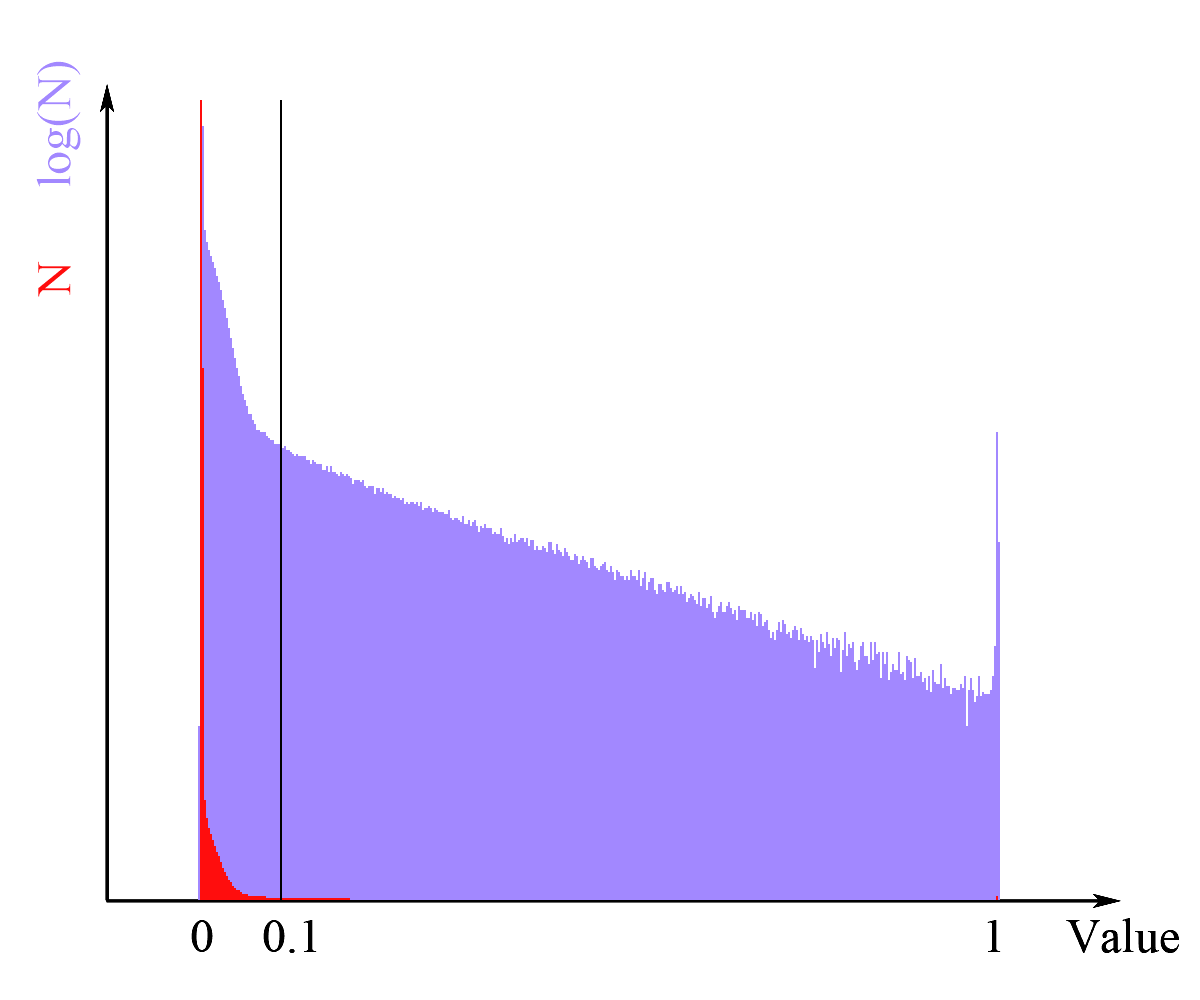}
\end{center}
\vspace{-1em}
\caption{
\zq{
\textbf{Distribution of the values in $\bm{T}$ after stage 1.} The distribution in logarithmic scale is shown in blue, and linear scale in red.
}
}
\label{fig:connection}
\end{figure}

%% file: fig/2Dtoy.tex
\begin{figure*}[t!]
\begin{center}
\includegraphics[width=1.0\linewidth]{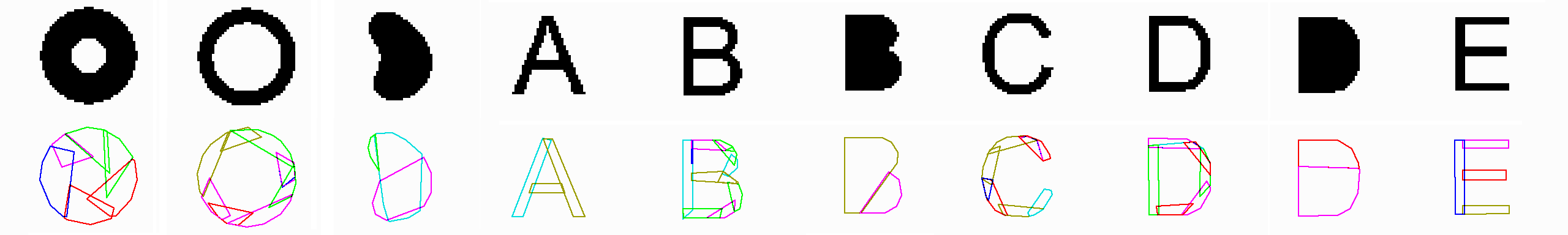}
\end{center}
\vspace{-.5em}
\caption{
\rz{
\textbf{Decomposing 2D shapes} using convexes learned by BSP-Net.
The resulting convexes are from autoencoders trained on some toy 2D shapes with overlap loss. Notice that in the visualization we use different (possibly repeating) colors to indicate different convexes.
}
}
\label{fig:2Dtoy}
\end{figure*}

%% file: tab/bspfc_numbers.tex
\begin{table}[t!]
\zq{
\caption{
\revision{Comparing general implicit field decoders on 3D Shape Auto-encoding}
\label{table:bspfc_numbers}
}
}
\vspace{-2em}
\begin{center}
\resizebox{1.0\linewidth}{!}{
\begin{tabular}{l|r|r|r}
\hline
 & CD & LFD & training time (h) \\
\hline\hline
DeepSDF					 & 0.534 & 2122.79 & 33.0   \\
IM-NET					 & 0.514 & 2029.87 & 62.9   \\
IM-NET*					 & 0.519 & 2050.40 & 47.8   \\
OccNet					 & {\bf 0.463} & {\bf 1899.75} & 92.6   \\
OccNet-/2				 & {\bf 0.468} & {\bf 1929.66} & 37.2   \\
DeepMETA				 & 0.611 & 2272.20 & {\bf 7.0}    \\
DeepMETA-x8				 & 0.590 & 2130.33 & {\bf 13.6}   \\
BSP-Net-2048p \revision{(stage 1)}  & 0.676 & 2438.00 & {\bf 15.0}   \\
BSP-Net-4096p \revision{(stage 1)}  & 0.685 & 2469.20 & {\bf 23.8}   \\
BSP-FC-4096p			 & 0.480 & {\bf 1958.60} & 44.0   \\
BSP-FC-2048p			 & 0.503 & 2010.87 & {\bf 26.6}   \\
BSP-FC-deeper-4096p		 & 0.481 & {\bf 1942.58} & 33.1   \\
BSP-FC-deeper-2048p		 & 0.492 & 2011.86 & {\bf 21.1}   \\
BSP-FC-QS-2048q		 & {\bf 0.471} & {\bf 1921.84} & {\bf 25.6}   \\
BSP-FC-QS-deeper-2048q & {\bf 0.475} & {\bf 1930.87} & {\bf 21.0}   \\
BSP-FC-QS-deeper-1024q & {\bf 0.477} & {\bf 1955.63} & {\bf 14.2}   \\
BSP-FC-QS-deeper-512q	 & 0.485 & {\bf 1994.08} & {\bf 12.0}   \\
\hline
\end{tabular}
}
\end{center}
\zq{
For CD, we highlight numbers below $0.48$; for LFD, we highlight numbers below 2,000; for training time, we highlight numbers below $30$.
}
\end{table}

%% file: fig/bspfc_tradeoff.tex
\begin{figure}[t!]
\begin{center}
\includegraphics[width=1.0\linewidth]{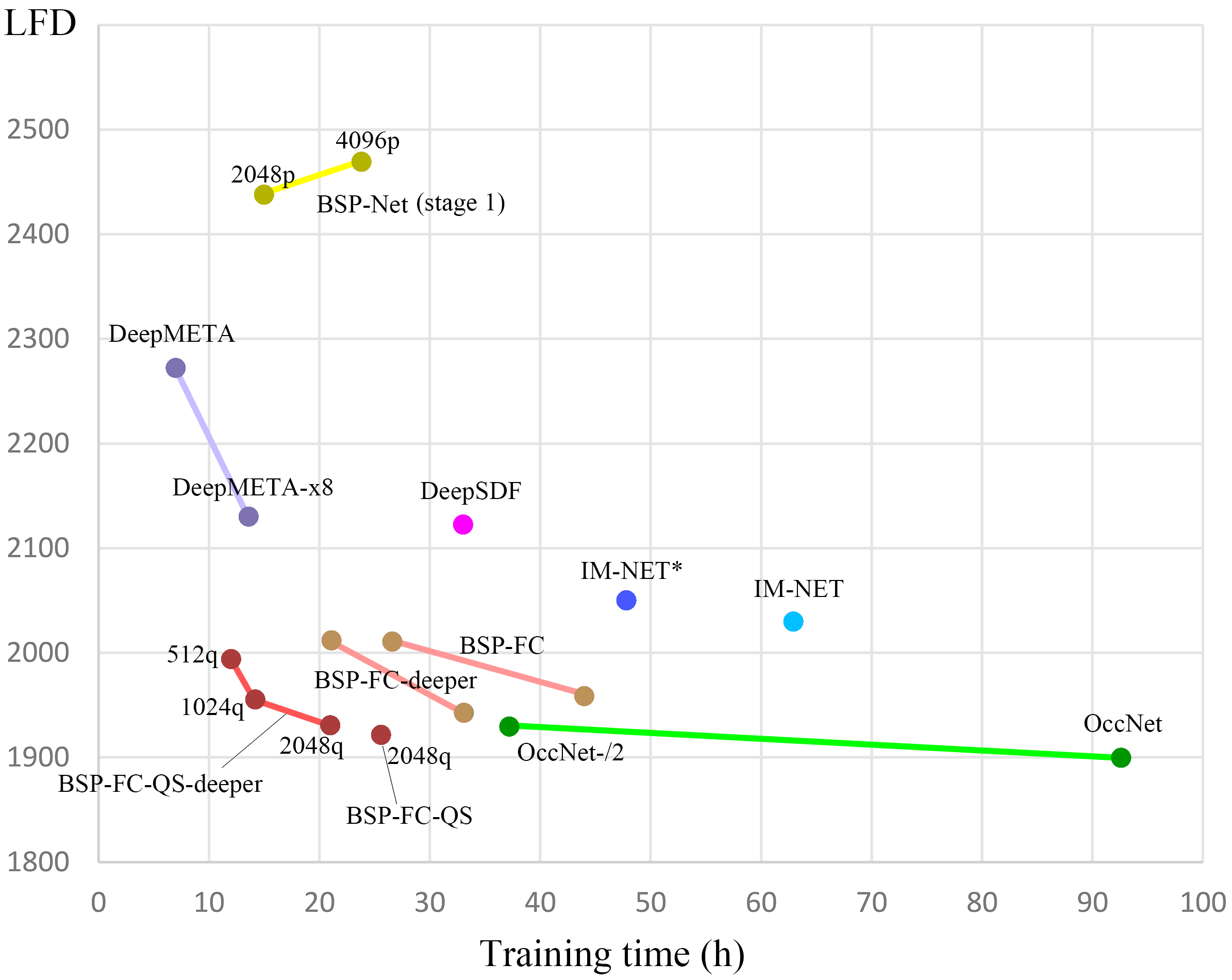}
\end{center}
\vspace{-.5em}
\caption{
\rz{
\textbf{Performance-training time trade-off} compared among different implicit decoder structures. Better trade-off is closer to the origin.
}
}
\label{fig:bspfc_tradeoff}
\end{figure}

%% file: fig/vae_interp.tex
\begin{figure*}[t!]
\begin{center}
\includegraphics[width=0.97\linewidth]{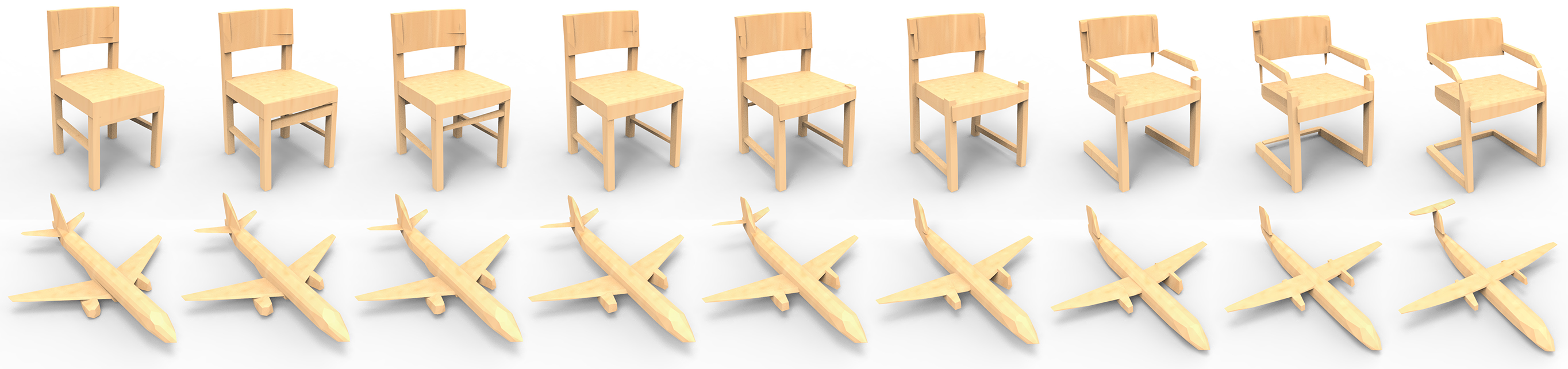}
\end{center}
\vspace{-.5em}
\caption{
\rz{\textbf{BSP-VAE shape interpolation.} The left- and right-most shapes are generated by an VAE, and the interpolation results are shown in the middle. During the interpolation, we linearly change the latent feature code, which is equivalent to linearly changing the plane parameters.}
}
\label{fig:vae_interp}
\end{figure*}

%% file: 5_future.tex
\section{Conclusion, limitation, and future work}
\label{sec:future}

We introduce BSP-Net, an unsupervised method which can generate compact and structured polygonal meshes in the form of convex decomposition.
Our network learns a BSP-tree built on the same set of planes, and in turn, the same set of convexes, to minimize a reconstruction loss for the training shapes.
These planes and convexes are defined by weights learned by the network.
Compared to state-of-the-art methods, meshes generated by BSP-Net exhibit superior visual quality, in particular, sharp geometric details, when a \textit{comparable} number of primitives are employed.

The main limitation of BSP-Net is that it can only decompose a shape as a \textit{union} of convexes. Concave shapes, e.g., a teacup or ring, have to be decomposed into many small convex pieces, which is unnatural and leads to wasting of a considerable amount of representation budget (planes and convexes); \rz{see \Figure{2Dtoy}.} A better way to represent such shapes is to do a \textit{difference} operation rather than union.
How to generalize BSP-Net to express a variety of CSG operations is an interesting direction for future work.

Current training times for BSP-Net are quite significant: 6 days for 4,096 planes and $256$ convexes for the SVR task trained across all categories; inference is fast however. While most shapes only need a small number of planes to  represent, we cannot reduce the total number of planes as they are needed to well represent a large {\em set\/} of shapes. 
It would be ideal if the network can adapt the primitive count based on the complexity of the input shapes; this may call for an architectural change to the network.

While its applicability to RGBD data could leverage the auto-decoder ideas explored by~\cite{DeepSDF}, the generalization of our method beyond curated datasets~\cite{chang2015shapenet}, and the ability to train from only RGB images are of critical importance.

%% file: 6_acks.tex
\ifCLASSOPTIONcompsoc
  
  \section*{Acknowledgments}
\else
  
  \section*{Acknowledgment}
\fi
We thank the anonymous reviewers for their insightful comments and the CVPR 2020 Best Paper Award Committee for their
recognition of our work.
This work was supported in part by an NSERC grant (No.~611370), a Google Faculty Research Award, and Google Cloud Platform research credits.